\documentclass[nohyperref]{article}

\usepackage{microtype}
\usepackage{graphicx}
\usepackage{subfigure}
\usepackage{booktabs} 

\usepackage[accepted]{icml2023}

\usepackage{amsmath}
\usepackage{amssymb}
\usepackage{mathtools}
\usepackage{amsthm}
\usepackage{multirow}
\usepackage{url}
\usepackage{algorithm}  
\usepackage{hyperref}
\usepackage{algorithmic}
\usepackage{bbm}
\usepackage{threeparttable}
\usepackage{xcolor}
\usepackage{wrapfig}
\usepackage[capitalize,noabbrev]{cleveref}
\usepackage[textsize=tiny]{todonotes}

\theoremstyle{plain}
\newtheorem{theorem}{Theorem}[section]

\newtheorem{lemma}[theorem]{Lemma}

\theoremstyle{definition}

\theoremstyle{remark}

\newcommand{\BenchName}{Hard-Bench }
\newcommand{\method}{Grad}
\newcommand{\methodRandom}{Random-Bench }
\newcommand{\methodgd}{Hard-Bench (GradNorm) }
\newcommand{\methodls}{Hard-Bench (Loss) }
\newcommand{\std}[1]{{\footnotesize \textcolor{gray}{$\pm$#1}}}
\newcommand{\gap}[1]{{\footnotesize #1}}
\newcommand{\badtb}[1]{\textbf{\textcolor{red}{#1}}}
\newcommand{\toptb}[1]{\textbf{\textcolor{blue}{#1}}}

\icmltitlerunning{A Challenging Benchmark for Low-Resource Learning}

\begin{document}

\twocolumn[
\icmltitle{A Challenging Benchmark for Low-Resource Learning}



\icmlsetsymbol{equal}{*}

\begin{icmlauthorlist}
\icmlauthor{Yudong Wang}{equal,comp,zzz}
\icmlauthor{Chang Ma}{equal,sch}
\icmlauthor{Qingxiu Dong}{yyy}
\icmlauthor{Lingpeng Kong}{sch}
\icmlauthor{Jingjing Xu}{comp}
\end{icmlauthorlist}

\icmlaffiliation{comp}{Shanghai Artificial Intelligence Laboratory}
\icmlaffiliation{yyy}{The MOE Key Laboratory of Computational Linguistics, Peking University}
\icmlaffiliation{sch}{The University of Hong Kong}
\icmlaffiliation{zzz}{Yingcai Honors College, University of Electronic Scienceand Technology of China}

\icmlcorrespondingauthor{Jingjing Xu}{jingjingxupku.02@gmail.com}

\icmlkeywords{Machine Learning, ICML}

\vskip 0.3in
]



\printAffiliationsAndNotice{\icmlEqualContribution} 

\begin{abstract}
With promising yet saturated results in high-resource settings, low-resource datasets have gradually become popular benchmarks for evaluating the learning ability of advanced neural networks (e.g., BigBench, superGLUE). Some models even surpass humans according to benchmark test results. However, we find that there exists a set of ``hard examples'' in low-resource settings that challenge neural networks but are not well evaluated, which causes over-estimated performance. 
We first give a theoretical analysis on which factors bring the difficulty of low-resource learning. It then motivates us to propose a challenging benchmark Hard-Bench to evaluate the learning ability better, which covers 11 datasets, including 3 computer vision (CV) datasets and 8 natural language process (NLP) datasets.
Experiments on a wide range of models show that neural networks, even pre-trained language models, have sharp performance drops on our benchmark, demonstrating the effectiveness on evaluating the weaknesses of neural networks.  On NLP tasks, we surprisingly find that despite better results on traditional low-resource benchmarks, pre-trained networks, does not show performance improvements on our benchmarks. These results demonstrate that there is still a large robustness gap between existing models and human-level performance. Code is available on \url{https://github.com/Qian2333/Hard-Bench}.  
\end{abstract}

\section{Introduction}

A common way to learn a promising model is to train neural networks on massive amounts of data for a certain downstream task, like machine translation~\citep{DBLP:journals/corr/BahdanauCB14}, object detection and recognition~\citep{DBLP:conf/cvpr/HeZRS16}, text-to-speech~\citep{DBLP:journals/corr/OordDZSVGKSK16}. As a comparison, neural networks still struggle to handle low-resource tasks that learn from a limited number of training samples. However, machine learning has recently been revolutionized by scaling up self-supervised pre-training, which confers various benefits, especially strong results in low-resource datasets~\citep{brown2020language, radford2021learning,bao2021beit}.  In this context, low-resource datasets have become prominent evaluation frameworks for research toward general-purpose learning abilities, such as superGLUE~\citep{DBLP:conf/nips/WangPNSMHLB19} and BigBench~\citep{DBLP:journals/corr/abs-2206-04615}. Furthermore, low-resource learning arises naturally in many real-world applications, such as personalized healthcare and low-resource language understanding, which makes low-resource benchmarks valuable.

\begin{figure}[t]
    \centering
    \includegraphics[width=\linewidth]{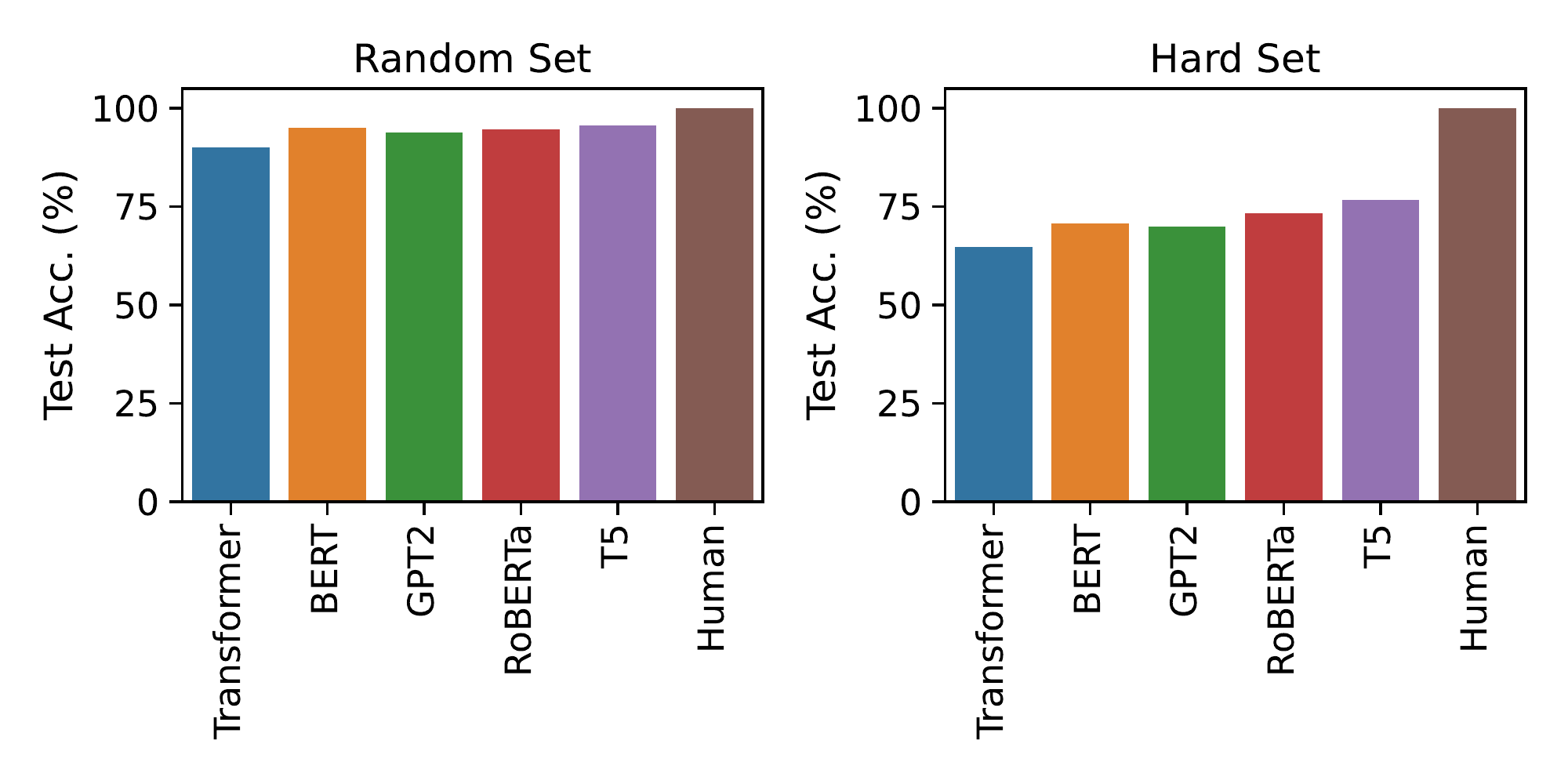}
    \caption{Results on sentiment classification (SST-2). The left figure shows average results on a randomly-sampled set as the test set. The right figure shows average results on a hard set as the test set. The hard test set is selected with smaller loss margins given a weak classifier. Although it is widely-accepted that neural networks can handle sentiment classification well with near-human accuracy (as shown in the left figure), the large drop on hard examples demonstrate that existing models still have generalization issues. }
    \label{fig:hard}
\end{figure}

However, current low-resource benchmarks are insufficient to evaluate the real learning gap between existing models and human-level models. On one hand, some models can surpass human performance according to benchmark accuracy~\citep{DBLP:conf/nips/YangDYCSL19,DBLP:conf/acl/NangiaB19,DBLP:conf/iclr/HeLGC21}. On the other hand, many previous studies have shown that these strong models still suffer from spurious correlation~\citep{DBLP:conf/icml/SagawaRKL20} or bias~\citep{DBLP:conf/nips/BolukbasiCZSK16}, which basically are rare in human learning. These inconsistencies demonstrate that current low-resource benchmarks lack difficulty and call for more challenging benchmarks. Also, current low-resource benchmarks are either randomly sampled from a training set or annotated on randomly-sampled data.  We give an example to show how randomly sampling a test set is insufficient to evaluate low-resource learning ability. Figure~\ref{fig:hard} shows results on sentiment classification, SST-2. The left part shows average results on a randomly-sampled set as the test set. The right part shows average results on a hard set as the test set. The hard test set is selected with smaller loss margins given a weak classifier. Random sampling usually contributes to unbiased distribution, which is relatively easier to handle. Yet humans are less sensitive than models to data bias and increased data difficulty, motivating us to propose a challenging low-resource benchmark.

In this work, we focus on low-resource settings and aim to build a challenging benchmark to better evaluate the learning ability. Unlike traditional low-resource datasets~\cite{dhillon2019baseline, ren2018meta, schick2020s} that aim to build a clean and unbiased low-resource datasets, we propose to include hard examples in our benchmark\footnote{To avoid involving mislabelled examples, we add the human-check process in this work.}. Since real-world low-resource data samples are more likely to be biased towards a certain domain (e.g. blank background for detection images, and short sentences for handwritten hate speech),  we also consider bias evaluation in this work. Specially, we consider two dimensions: misleading examples with smaller classification margins for performance evaluation and biased examples for robust evaluation. We first give a comprehensive analysis on how these two dimensions affect low-resource learning and introduce an empirical solution to build a challenging low-resource benchmark based on our analysis results. The final benchmark covers 3 CV datasets and 8 NLP datasets. 

To prove the effectiveness of the constructed benchmark, we evaluate 11 models, including 7 pre-trained models. All these models struggle on handling our benchmarks, with a large performance gap compared with randomly-sampled low-resource benchmarks. On NLP tasks, we surprisingly find that despite with better results on traditional low-resource benchmarks, pre-trained networks, does not show performance improvements on our benchmarks. These results prove that there are still a large performance gap between existing models and human-level performance. The contribution of this paper is summarized as:
\begin{itemize}
    \item We propose a challenging benchmark Hard-Bench towards weaknesses of neural networks.
    
    \item We give a comprehensive analysis on which factors affect the difficulty of low-resource learning.

    \item Experiments show that Hard-Bench can better tell the learning gap between existing models compared with randomly-sampled low-resource datasets.
\end{itemize}

\section{Related Work}

\textbf{Low-resource Evaluation}
Learning on low-resource datasets has recently come into the spotlight with the introduction of more powerful models ~\citep{brown2020language, radford2019language} and benchmarks on low-resource evaluation are ever-catching the footsteps of model advancement to challenge SOTAs. Traditional few-shot learning benchmarks, i.e. miniImageNet~\citep{cai2018memory}, CIFAR-FS~\citep{bertinetto2018meta}, focus on the meta-learning N-way K-shot setting, but this setting is less applicable to transfer learning models~\citep{dumoulin2021unified} and was demonstrated not challenging enough for large pre-trained models~\citep{dhillon2019baseline}. Recent low-resource benchmarks have expanded their settings to transfer learning scenarios~\citep{dumoulin2021unified,zheng2021fewnlu} as well as in-context learning~\citep{bragg2021flex, schick2020s}, and they have also added up on dataset difficulty~\citep{wang2018glue}. Among these, there are two major types of low-resource benchmark: natural low-resource datasets, and sampled low-resource datasets. The former requires additional dataset curation~\citep{wang2018glue, DBLP:journals/corr/abs-2206-04615, koh2021wilds} and currently, most low-resource benchmarks are uniformly sampled from larger datasets~\citep{kolesnikov2020big, schick2020s, alayrac2022flamingo, logan2021cutting, brown2020language}. We take a step further to review the process they use to sample low-resource datasets and propose a simple yet effective method to challenge current models.

\textbf{Data Pruning} Our approach is similar to data pruning literature in that we  both hope to find a difficult subset in a large dataset. Previously, data pruning methods \citep{hacohen2019power, paul2021deep, toneva2018empirical, sorscher2022beyond} use data difficulty metrics including GradNorm and Loss Score to rank and prune datasets. However, we approach dataset sampling from a drastically different goal as we hope to challenge low-resource learning models.  

\section{Understanding the Difficulty of Low-Resource Learning\label{section: low resource analysis}}

\begin{figure}
    \centering
    \includegraphics[width=0.9\linewidth]{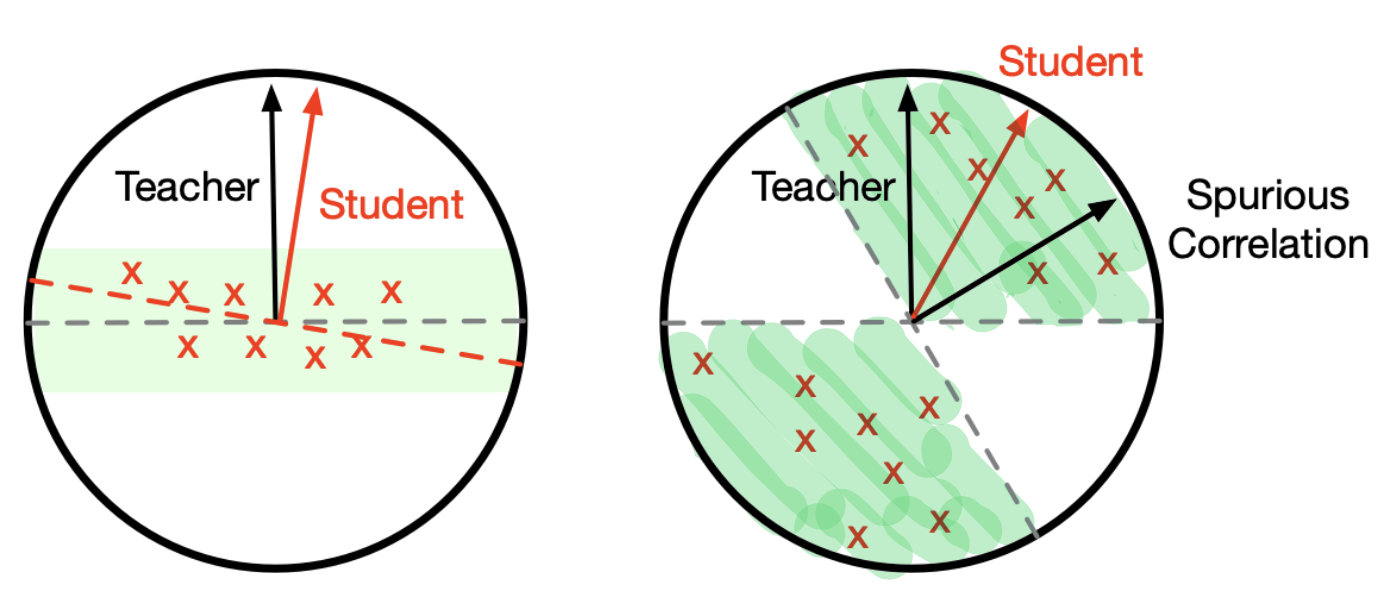}
    \caption{Plot of the perceptron model under hard low-resource learning (left) and biased low-resource learning setting (right). The green area shows the region where few-shot samples are sampled. (a) Under the hard low-resource learning setting, data samples are selected within a small margin to the decision boundary. (b) Under the biased low-resource learning setting, data samples are selected to satisfy the spurious classifier.}
    \label{fig:analysis main}
\end{figure}

To better understand the low resource scenario in deep learning, we first look at a the teacher-student setting in learning perceptrons. Consider a large curated dataset of $N$ examples $D=\{x_i,y_i\}_{i\in [N]}$ where $x_i\in \mathbbm{R}^d$ are i.i.d. random Gaussian inputs $x_i \sim \mathcal{N}(0, I_d)$, with labels generated by a teacher perceptron $T\in \mathbbm{R}^d$ as $y_i = \text{sign}(Tx_i)$. The number of samples $N\rightarrow \infty$ but sample per parameter $\alpha = \frac{N}{d}=O(1)$ to remain trainable. Now we consider the low resource scenario where the number of training samples available $P$ is much less than $N$, where $\alpha_{\text{low}}=\frac{P}{d}\rightarrow 0$. For convenience, we sample the data for low resource learning from dataset $D$ such that $D_{\text{low}}=\{x_{\mu},y_{\mu}\}_{i\in[P]}\subset D$. Learning on $D_{\text{low}}$, we obtain a new student perceptron $J$ that has generalization error $\epsilon_g$. 

Intuitively, three dimensions amount to the difficulty of learning perceptron $J$: (1) the number of training samples $P$ (here we base the study of data scarity on the sample per parameter variable $\alpha_{\text{low}}$); (2) the classification difficulty of the data samples, denoted by the margin $m = \min_{\mu}J(x_\mu y_\mu)$; (3) the  bias of the training dataset:  here we look at a specific type of bias, spurious correlation, which draws correlation based on peripheral attributes of data items with a target variable, denoted as a student perceptron $J_{\text{bias}}$. We explore the difficulty of low-resource learning by altering our selection procedure for $D_{\text{low}}$ and explore how $\epsilon_g$ changes. Specifically, we look at three settings and use simulation experiments for analysis.
\begin{itemize}
    \item Low-resource learning, where $D_{\text{low}}$ is uniformly sampled from $D$.
    \item Hard low-resource learning, where the margin of each sample is calculated $m_\mu = T(x_\mu y_\mu)$ and the samples with the smallest margins are selected from $D$, as shown in Figure \ref{fig:analysis main}.
    \item Biased low-resource learning, where a biased probe $J_{\text{bias}}$ with $\theta$ angle to $T$ is chosen as the spurious classifier. Then data that satisfies both $y_i = \text{sign}(J_{\text{bias}}x_i)$ and $y_i = \text{sign}(Tx_i)$ is uniformly sampled from $D$, as shown in Figure \ref{fig:analysis main}.
\end{itemize}

We elaborate on simulation settings in the Appendix \ref{appendix: perceptron theory}.

\begin{figure}[t]
    \centering
    \includegraphics[width=0.5\linewidth]{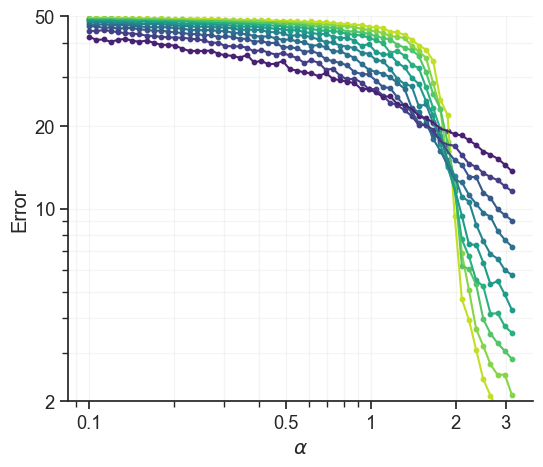}
    \caption{Plot of generalization error with regard to the number of samples per parameter. Darker lines represent easier data samples.}
    \label{fig:error_hard}
\end{figure}

\textbf{Difficult data especially challenges low resource learning.} We first compare the setting that increases data difficulty to the random-sampled version of Low-resource Learning. We vary our dataset size from 1\% to 500\% trainable parameters. As shown in Figure \ref{fig:error_hard}, the dark blue line corresponds to the setting where data is uniformly selected, and lighter lines range in data difficulty from margin 0.1 to 1. The functions of $\epsilon_g$ to $\alpha$ yield a crossover between the function for random-sampled training data and the one for increased difficulty training data, showing that increased data difficulty affects low resource settings more than sufficient data settings. Also, the increase in generalization error is more distinct for slightly larger training sets. As when the low-resource training set only has a few samples, it requires model to have strong generalization ability to beat the rule of generalization $\epsilon \propto \alpha^{-1}$ and the task is challenging enough.

\begin{figure}[htbp]
    \centering
    \resizebox{\linewidth}{!}{
\includegraphics{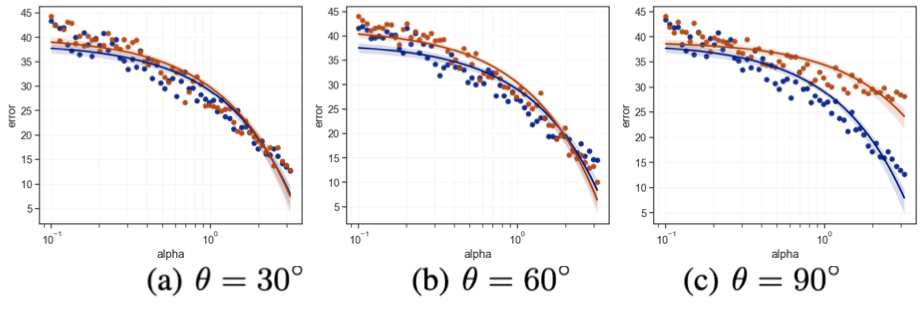}
}
    \caption{Plot of generalization error with regard to the number of samples per parameter. Red lines represent biased training set, while blue lines represent unbiased training set.}
    \label{fig: analyze bias}
\end{figure}

\textbf{Low resource learning is more sensitive to spurious correlations.} In the biased learning scenario as shown in Figure~\ref{fig: analyze bias}, we compare students trained on biased datasets (red lines) to students trained on random-sampled datasets (blue lines). When the bias probe is more distinct from the teacher (larger $\theta$), the drop in performance is more distinct. This is in line with the phenomenon that when a model overfits on spurious features that contain information distant from semantics, the model tends to suffer on generalization. Also, for smaller bias, low resource learning sees a larger drop in generalization while models with abundant data barely suffer. This show that low-resource learning is sensitive to even small biases. 

\textbf{Theoretical perspective} Here we use theoretical analysis in addition to simulations to study the scenario that results in failed generalization in low resource learning. Again, we focus on the scenario where we have a large dataset $D$ that represents the natural task distribution $P$. We sample a low resource dataset $D_{low}$ from $D$ that form the distribution $P_{low}$. We theoretically show that the generalization error on test set $Q$ for the model trained on the low-resource dataset is bounded by a function of data difficulty and the distribution bias of low-resource dataset. 

\begin{theorem}\label{covariate shift}\textit{(Low-resource Generalization Measured by Distribution Shift and Data difficulty) } Let $\mathcal{H}$ be the hypothesis space $X\rightarrow \mathbbm{R^d}$. $f_{low}$ is the empirical risk $\epsilon_{P_{low}}(f)$ minimizer, and $f$ is the hypothesis that minimizes expected risk $\epsilon_{Q}(f)$, m is the smallest margin of $D$ to decision boundary of $f$. Then with probability over 1-$\delta$, 
    \begin{equation}
    \label{equation: main theorem}
    \begin{aligned}
    \epsilon_Q(f_{low}) \leq \epsilon_Q(f)+c \sqrt{\frac{|\mathcal{H}|\ln m+\ln \left(\frac{2}{\delta}\right)}{m}} \\ +\text{MMD}(P_{low}, P) +  \epsilon_{\alpha} + \epsilon_{\mathcal{H}}
    \end{aligned}
\end{equation}
where $\epsilon_{\alpha}$, and $\epsilon_{H}$ are small constants describing the error occurred in training, the sampling behavior of training distribution and the hypothesis space complexity. Details are shown in Appendix~\ref{section: proof theorem}.
\end{theorem}

The value of the Equation~\ref{equation: main theorem} right-hand side increases when $m$ decreases and the term $\text{MMD}(P_{low}, P)$ increases, corresponding to the increase in data difficulty and the presence of data bias. Note that this theorem applies not only to our simulated scenario of perceptron learning but also to deeper models. In our biased low-resource learning setting, the distribution gap between low resource data distribution is larger for biased training set than random-sampled training set, i.e., $\text{MMD}(P^{\theta}_{low}, P) > \text{MMD}(P^{\text{random}}_{low}, P)$, since data samples forming $P_{low}^{random}$ are sampled uniformly from $P$.

Based on our simulation experiments and theoretical results in the previous section, we find that low-resource learning is more likely to suffer from performance drop due to data difficulty and dataset bias. However, these scenarios are not covered in previous low-resource benchmarks. This motivates us to propose a challenging benchmark hardBench for better evaluation. 

\section{Hard-Bench Challenge}
We propose a new challenging benchmark that elevates low-resource learning difficulty on some well-known datasets. Unlike previous low-resource datasets that are randomly sampled from a training set, we curate the benchmark by selecting one of the most challenging low-resource training sets from CIFAR10, CIFAR100, ImageNet, and GLUE. In this section, we focus on the metrics for dataset selection. 

Following our theoretical analysis, we introduce the simple yet effective approach to build hardBench: First, we train a predictor for only one epoch on a large benchmark, obtaining a biased predictor; then, we score each sample on data difficulty for this stage of training. For each label, we pick the top k samples as our selected low-resource training set. We elaborate on the data difficulty metrics and the biased predictor respectively in section~\ref{section: difficulty metrics}. 

\subsection{Metrics Measuring Data Difficulty \label{section: difficulty metrics}}

Previous literature in curriculum learning~\citep{hacohen2019power}, data pruning~\citep{paul2021deep}, and continual learning~\citep{toneva2018empirical} propose metrics for data sample difficulty based on loss or gradient norms. Here we restate three metrics: \textit{Loss score}, \textit{GradNorm score} and explain how they can be applied in our problem scenario. 

\textbf{Loss Score}
\citet{paul2021deep} and \citet{sorscher2022beyond} state this metric in the EL2N method, which intuitively measure data samples difficulty by looking at whether they can be learned correctly. Data samples with a higher loss score after training are more likely to be near the decision boundary. Therefore, we can select the hardest samples by ranking the loss score on the dataset. We call datasets constructed via loss scores as \textbf{Hard-Bench (Loss)}. Examples with higher losses are selected as hard examples. 

\textbf{Gradient Norm Score} \citet{paul2021deep} discussed using gradient norm as an indicator of data importance. Samples with larger gradient norms shape the training geometry. However, there is little discussion on the connection between gradient norm and data difficulty. Here we give a brief and casual explanation. Based on previous analysis, we can find hard samples by checking their margin to the decision boundary of our model $f$, $f(x_0)=0$. Therefore, we can define the $L_p$ norm margin as, 
\begin{equation}
    m(x) = \min_{x_0} ||x - x_0||_p, s.t. f(x_0)=0 
\end{equation}
We can use Taylor's approximation to find an approximate solution to the problem, following~\citep{elsayed2018large}. 
\begin{equation}
\begin{aligned}
m(x) & =\min _r\|r\|_p \quad \text { s.t. } \quad f(x+r)=0 \\
& \approx \min _r\|r\|_p \quad \text { s.t. } \quad f(x)+\nabla_x f(x)^T r=0 \\
& =\frac{|f(x)|}{\left\|\nabla_x f(x)\right\|_q},
\end{aligned}
\end{equation}

When the numerator is constrained (For a classification problem, we can constraint logits $f(x)$ within 1 using sigmoid function. ), we can maximize the gradient norm to minimize margin. We call datasets constructed via gradient norm scores as \textbf{Hard-Bench (GradNorm)}. Examples with higher gradient norm scores are selected as hard examples. 

\begin{figure}[t]
    \centering
    \includegraphics[scale=0.3]{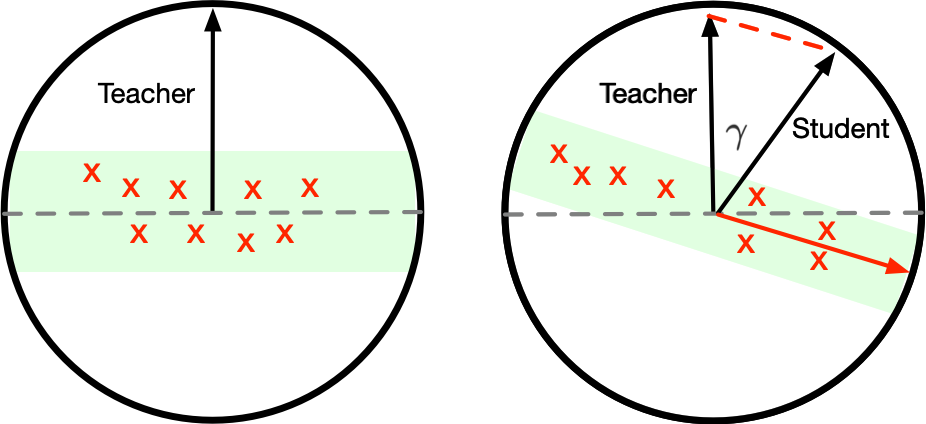}
    \caption{Plot of the perceptron model under both hard and biased low-resource learning setting. Compared to the no-bias setting on the left, when the gap between $J_{bias}$ and teacher is $\gamma$, the resulting bias of dataset would be $\frac{\gamma}{2}$.}
    \label{fig:analysis bias}
\end{figure}


\subsection{Introducing Bias with Early Stopping\label{section: biased classifier}}

As shown in the above sections, we need to train a student predictor to estimate the decision boundary and thereby calculate the data difficulty score. However, we find that we can easily introduce bias into our selected benchmark dataset if we early stop training on the student predictor. We will give an explanation based on the Loss Score.

The Loss Score effectively estimates the difficulty of data examples to be classified correctly when the student predictor is exactly the same as the teacher model, i.e. $\theta = 0 $. However, when the student model is undertrained, there would exist a gap $\gamma$ between student $g(x)=sgn(Jx)$ and teacher $f(x)=sgn(Tx)$. For any $x$, the loss function would be $L(x) = g(x)-f(x) = (J-T)x$. Therefore, the resulting selected dataset $D_{low}=\{(x_i,y_i)| x_i=\max^{i=1,2,...P}_{x} (J-T)x, y_i = \text{sgn}(Tx_i)\}$ is isotropic in the nullspace of $J-T$, inducing a bias of $\frac{\gamma}{2}$. 

This intuitively explains that we can use an early stopped predictor as well as data difficulty metrics to select a biased and difficult low-resource dataset that mimics the real-world setting. In the following sections, we use this approach to curate our Blindsight Benchmark and benchmark current models.





\section{Experiments}


 \begin{table*}[t!]
	\centering
	\caption{
		 Results on NLP datasets with 500 shots. Hard-Bench (Loss) brings higher performance drops than Hard-Bench (GradNorm). Surprisingly, pre-trained networks does not show better generalization results than randomly-initialized models on our benchmark.
	}
    \footnotesize
	\resizebox{0.9\textwidth}{!}{%
		\begin{tabular}{lccccccccc}
			\toprule
   
 \multirow{1}{*}{ \textbf{Models}} & \multirow{1}{*}{\textbf{SST2}} & \multirow{1}{*}{\textbf{COLA}} & \multirow{1}{*}{\textbf{MNLI}} & \multirow{1}{*}{\textbf{QNLI}} & \multirow{1}{*}{\textbf{MRPC}} & \multirow{1}{*}{\textbf{QQP}} & \multirow{1}{*}{\textbf{RTE}} & \multirow{1}{*}{\textbf{WNLI}}& \multirow{1}{*}{\textbf{Average}}\\

			\midrule
 \multicolumn{9}{l}{\textit{\methodRandom}} \\
			\midrule
      
 Transformer&  {68.16}\std{1.46} & {69.15}\std{0.04} &  {36.42}\std{0.58} &  {55.45}\std{0.94} &  {68.58}\std{0.39} &  {67.18}\std{0.67} &  {53.65}\std{1.04} &  56.34\std{0.00} &  59.37\\
 BERT       &  88.68\std{0.73} &  79.00\std{0.59} &  57.60\std{1.30} &  76.02\std{1.26} &  77.65\std{1.38} &  75.53\std{0.48} &  60.58\std{2.01} &  \badtb{48.45}\std{3.63} &  70.44\\
 GPT-2      &  88.08\std{0.72} &  70.35\std{1.76} &  58.35\std{1.65} &  74.11\std{2.56} &  75.93\std{0.47} &  76.22\std{0.86} &  65.49\std{2.62} &  \toptb{56.90}\std{3.40}&  70.68\\
 RoBERTa    &  \toptb{91.54}\std{0.61} &  \toptb{80.98}\std{0.56} &  \toptb{75.4}0\std{0.52} &  \toptb{84.47}\std{0.53} &  \toptb{88.24}\std{0.27} &  \toptb{80.93}\std{0.56} &  \toptb{73.00}\std{1.98} &  54.93\std{2.82}&  \toptb{78.69}\\
 T5         &  88.73\std{0.97} &  78.62\std{0.58} &  64.53\std{2.48} &  82.56\std{0.83} &  74.56\std{1.71} &  80.13\std{0.44} &  56.46\std{1.95} &  52.39\std{7.74} & 72.25\\

			\midrule
 \multicolumn{9}{l}{\textit{\methodgd}} \\
			\midrule
      
 Transformer&  51.88\std{0.46} &  {69.15}\std{0.04} &  35.11\std{0.67} & {50.59}\std{0.04} &  68.38\std{0.00} &  {62.41}\std{1.06} &  54.01\std{0.96} &  {56.34}\std{0.00} &  55.98\\
 BERT       &  47.94\std{2.11} &  45.77\std{8.19} &  33.96\std{0.47} &  46.24\std{2.35} &  56.08\std{1.43} &  52.60\std{3.01} &  51.12\std{0.96} &   {49.30}\std{1.99} &  47.88\\
 GPT-2      &  51.44\std{0.77} &   {51.93}\std{7.92} &  35.98\std{1.95} &   {48.62}\std{5.12} &  65.98\std{2.33} &  55.40\std{4.05} &  57.76\std{4.60} &  56.06\std{2.25} &  52.90\\
 RoBERTa    &  51.01\std{0.65} &  66.10\std{6.01} &  {38.42}\std{1.51} &  48.61\std{1.50} &  {82.55}\std{1.04} &  56.69\std{3.93} & {60.36}\std{2.88} &   {54.93}\std{2.18} &  57.33\\
 T5         &  {52.34}\std{2.35} &   {55.09}\std{6.16} &  34.27\std{0.39} &  48.99\std{1.51} &  55.88\std{3.80} &  55.72\std{1.62} &  48.88\std{1.54} &  54.37\std{4.14} &  50.69\\

			\midrule
 \multicolumn{9}{l}{\textit{\methodls}} \\
			\midrule
      
Transformer&   {51.38}\std{0.40} &   {69.11}\std{0.04} &  {34.98}\std{0.69} &   {50.57}\std{0.04} &   {65.64}\std{5.49} &   {48.17}\std{7.69} &   {53.43}\std{0.40} &   {56.34}\std{0.00} &  53.70\\
 BERT       & \badtb{45.64}\std{5.32} & \badtb{40.92}\std{4.29} & \badtb{30.55}\std{0.88} & \badtb{40.11}\std{3.69} &   {38.24}\std{2.52} &   {35.55}\std{2.57} &  \badtb{ 47.44}\std{1.22} &  53.52\std{3.67} &  \badtb{41.50}\\
 GPT-2      &   {49.79}\std{2.06} &  56.18\std{9.92} &   {31.41}\std{1.19} &  {51.01}\std{3.89} &  50.54\std{8.02} &   {40.33}\std{5.57} &   {54.73}\std{3.67} &   {55.49}\std{1.44} &  48.69\\
 RoBERTa    &   {50.55}\std{0.62} &   {48.32}\std{11.78} &   {31.66}\std{2.49} &   {41.79}\std{5.62} &   \badtb{38.14}\std{2.54} &   \badtb{31.74}\std{2.44} &   {55.09}\std{1.97} &  55.77\std{1.91} &  44.13\\
 T5         &   {49.86}\std{2.85} &  55.32\std{6.06} &   {32.76}\std{0.23} &   {47.15}\std{1.76} &         
  {53.19}\std{5.12} &   {48.84}\std{5.38} &   {48.45}\std{1.20} &   {53.52}\std{4.45} &  48.64\\

\bottomrule
		\end{tabular}%
	 }
	\label{tab:02_main_nlp1}
\end{table*}


\begin{table*}[t!]
	\centering
	\caption{
		 Results on NLP datasets with 16 shots. BLOOM adopt ICL for few-shot inference.
	}
    \footnotesize
	\resizebox{0.9\textwidth}{!}{%
		\begin{tabular}{lccccccccc}
			\toprule
   
 \multirow{1}{*}{ \textbf{Models}} & \multirow{1}{*}{\textbf{SST2}} & \multirow{1}{*}{\textbf{COLA}} & \multirow{1}{*}{\textbf{MNLI}} & \multirow{1}{*}{\textbf{QNLI}} & \multirow{1}{*}{\textbf{MRPC}} & \multirow{1}{*}{\textbf{QQP}} & \multirow{1}{*}{\textbf{RTE}} & \multirow{1}{*}{\textbf{WNLI}}& \multirow{1}{*}{\textbf{Average}}\\

			\midrule
 \multicolumn{9}{l}{\textit{\methodRandom}} \\
			\midrule
      
 Transformer&  {52.64}\std{2.16} & {68.95}\std{0.44} &  {35.40}\std{0.09} &  {53.48}\std{2.46} &  {68.63}\std{0.31} &  {63.75}\std{0.55} &  {53.72}\std{0.90} &  58.03\std{2.07} & 56.82 \\
 BERT       &  68.39\std{7.14} &  66.94\std{3.55} &  36.21\std{0.96} &  53.75\std{0.69} &  66.47\std{3.22} &  64.81\std{2.15} &  55.02\std{1.56} &  56.34\std{1.56} & 58.49  \\
 GPT-2      &  55.62\std{4.12} &  66.40\std{5.50} &  37.63\std{1.29} &  55.16\std{3.26} &  67.75\std{1.53} &  62.57\std{1.34} &  \toptb{58.77}\std{3.98} &  56.34\std{0.00}& 57.53 \\
 RoBERTa    &  \toptb{76.67}\std{3.44} &  \toptb{69.66}\std{1.02} &  \toptb{43.13}\std{2.07} &  \toptb{63.52}\std{3.92} &  \toptb{69.26}\std{1.48} &  \toptb{65.55}\std{1.36} &  55.16\std{1.73} &  57.75\std{3.09}&  \toptb{62.58} \\
 T5         &  55.94\std{3.74} &  55.82\std{8.92} & 33.98\std{0.50} &  54.03\std{2.36} &  58.58\std{5.94} &  55.49\std{3.35} &  51.05\std{2.44} &  \toptb{58.31}\std{0.69} & 52.90\\
 \midrule
  BLOOM-7.5B &  50.46 \std{0.00} &  
60.40 \std{0.00} & 35.42 \std{0.00} & 50.54 \std{0.00} &  66.18 \std{0.00}  & 51.79 \std{0.00} &  52.70 \std{0.00} & 42.25 \std{0.00} & 51.21 \\

			\midrule
 \multicolumn{9}{l}{\textit{\methodgd}} \\
			\midrule
      
 Transformer&  52.89\std{0.56} &  {68.74}\std{0.86} &  35.45\std{0.00} & {50.95}\std{0.55} &  68.38\std{0.00} &  {63.23}\std{0.08} &  54.95\std{1.01} &  {56.62}\std{0.56} & 56.40 \\
 BERT       &  56.86\std{4.88} &  64.99\std{6.39} &  34.21\std{0.54} &  50.79\std{0.31} &  63.19\std{6.52} &  59.19\std{2.86} &  53.43\std{2.41} &   {54.37}\std{4.51} & 54.62 \\
 GPT-2      &  52.52\std{2.00} &   {66.19}\std{5.92} &  34.40\std{1.39} &   {53.65}\std{2.94} &  66.23\std{4.44} &  54.64\std{4.89} &  58.84\std{2.84} &  56.62\std{1.38} &  55.38 \\
 RoBERTa    &  58.12\std{1.47} &  65.23\std{5.14} &  {35.48}\std{0.82} &  50.80\std{0.37} &  {57.60}\std{7.79} &  63.18\std{0.00} & {53.14}\std{0.42} &   {56.90}\std{1.13} & 55.05 \\
 T5         &  {51.95}\std{1.90} &   {59.54}\std{4.84} &  33.44\std{0.21} &  50.85\std{1.02} &  59.90\std{4.06} &  56.61\std{3.44} &  49.03\std{1.76} &  53.52\std{5.12} & 51.85 \\
 \midrule
  BLOOM-7.5B &  51.15 \std{0.00} &  
33.34 \std{0.00} & 35.33 \std{0.00} & 50.47 \std{0.00} & 65.44\std{0.00} & 60.78 \std{0.00} & 47.29 \std{0.00} & 42.25 \std{0.00} & 48.25 \\

			\midrule
 \multicolumn{9}{l}{\textit{\methodls}} \\
			\midrule
      
Transformer&   {52.34}\std{0.26} &   {68.88}\std{0.50} &  {35.28}\std{0.21} &   {51.22}\std{0.38} &   {68.33}\std{0.10} &   {63.19}\std{0.02} &   {54.80}\std{0.42} &   {57.46}\std{1.05} & 56.43 \\
 BERT       & \badtb{50.28}\std{0.88} & 58.16\std{13.44} & 34.03\std{0.96} & 50.10\std{0.66} &   {54.61}\std{16.68} &   {57.27}\std{4.69} &  50.40\std{2.59} &  56.62\std{2.87} &  51.43 \\
 GPT-2      &   {51.54}\std{1.22} &  66.56\std{5.19} &   {33.88}\std{1.31} &  {52.49}\std{2.08} &  63.87\std{8.90} &   {56.84}\std{3.40} &   {52.71}\std{1.69} &   {56.34}\std{1.54} & 54.27 \\
 RoBERTa    &   {50.25}\std{0.96} &   \badtb{49.38}\std{11.73} &   {33.38}\std{1.10} &   {49.78}\std{0.40} &   \badtb{33.33}\std{0.83} &   63.10\std{0.10} &   {52.56}\std{0.37} &  56.34\std{1.64} & \badtb{48.51} \\
 T5         &   {51.19}\std{2.26} &  56.80\std{6.79} &   \badtb{33.39}\std{0.18} &   \badtb{49.69}\std{0.94} &         
  {56.32}\std{8.15} &   \badtb{56.14}\std{2.58} &   \badtb{49.10}\std{0.97} &   \badtb{52.11}\std{5.42} & 50.59 \\
   \midrule
    BLOOM-7.5B &  50.11 \std{0.00} &  
46.40 \std{0.00} & 35.42 \std{0.00} &50.01 \std{0.00} &65.93 \std{0.00} &60.79 \std{0.00} &47.29 \std{0.00} &43.66 \std{0.00} & 49.95 \\

\bottomrule
		\end{tabular}%
	 }
	\label{tab:02_main_nlp16shot}
\end{table*}

\begin{table}[t]
\centering
\caption{
    Results on CV datasets. Hard-Bench (Loss) and Hard-Bench (GradNorm) challenge neural networks with large performance drops, especially for random-initialized networks. ViT and efficientNetV2-S are pre-trained on ImageNet, and we do not report their results on ImageNet to avoid data leak issues. 
}
     \footnotesize
 \resizebox{0.85\linewidth}{!}{%
		\begin{tabular}{lccc}
			\toprule
\textbf{Models} & \textbf{CIFAR10} & \textbf{CIFAR100} & \textbf{ImageNet} \\

\midrule
 \multicolumn{4}{l}{\textit{\methodRandom}} \\
\midrule
FFN              &  48.91\std{0.87} &  14.95\std{0.29} &   5.12\std{0.30} \\
VGG-16           &  62.15\std{0.71} &  26.55\std{0.20} &  16.02\std{0.27} \\
ResNet-18        &  65.47\std{0.84} &  25.49\std{0.60} &  29.34\std{0.31} \\
DenseNet-121     &  71.33\std{0.56} &  33.66\std{1.48} &  \toptb{35.20}\std{0.41} \\
ViT-B/16         &  97.20\std{0.22} &  \toptb{83.93}\std{0.43} &   -\\
EfficientNetV2-S &  91.41\std{0.60} &  70.41\std{0.74} &   -\\

\midrule
 \multicolumn{4}{l}{\textit{\methodgd}} \\
\midrule

FFN              &  29.64\std{0.88} &   8.75\std{0.28} &   3.13\std{0.18} \\
VGG-16           &  55.11\std{0.89} &  17.22\std{0.44} &   9.51\std{0.20} \\
ResNet-18        &  46.87\std{2.41} &  15.50\std{0.85} &  23.81\std{0.76} \\
DenseNet-121     &  59.87\std{0.66} &  20.96\std{0.94} &  28.96\std{0.67} \\
ViT-B/16         &  \toptb{97.39}\std{0.10} &  82.36\std{0.94} &  - \\
EfficientNetV2-S &  92.51\std{0.24} &  69.56\std{0.49} &  - \\
 
\midrule
\multicolumn{4}{l}{\textit{\methodls}} \\
\midrule

FFN              & \badtb{17.26}\std{0.82} & \badtb{3.18}\std{0.21} & \badtb{2.66}\std{0.02} \\
VGG-16           &   {27.58}\std{0.62} &   { 7.14}\std{0.24} &   { 7.27}\std{0.24} \\
ResNet-18        &   {33.20}\std{1.00} &   { 6.96}\std{0.32} &   {13.34}\std{0.19} \\
DenseNet-121     &   {44.81}\std{2.30} &   {11.59}\std{0.98} &   {22.00}\std{0.46} \\
ViT-B/16         &   {96.85}\std{0.11} &   {80.87}\std{0.58} &  -\\
EfficientNetV2-S &   {89.88}\std{0.63} &   {60.42}\std{1.85} & - \\
			\bottomrule
		\end{tabular}
}
	\label{tab:01_cv_main}
\end{table}

In this section, we evaluate 11 models, including 7 pre-trained models, on our benchmark to see the performance of strong models on handling ``hard examples''.

\subsection{Benchmark Metric}

Since traditional low-resource benchmarks usually randomly choose a subset from the full-size training data as the training set. In this paper, we also follow this setting and extract hard examples from the full-size data as the training data in our benchmark. To be specific, we implement three benchmarks in this work, which are described as follows.
\begin{itemize}
    \item \textbf{Random-Bench}. For each label, we randomly select $k$ examples as the training set. We randomly select 3 subsets and report the average results. 
    \item \textbf{Hard-Bench (Loss)}. For each label, we choose top-$k$ hard examples based on losses scores. 
    \item \textbf{Hard-Bench (GradNorm)}.  For each label, we choose top-$k$ hard examples based on gradient norm scores. 
\end{itemize}

\subsection{Benchmark Datasets}

\paragraph{CV} We explore 3 widely-used image classification datasets, CIFAR-10~\citep{cifar}, CIFAR-100~\citep{cifar}, and ILSVRC-2012 ImageNet~\citep{deng2009imagenet}. We select a subset of the full-size training set as the training set in our benchmark where each label keeps $k$ examples. For all CV datasets, we implement a simple feed-forward networks trained with one epoch as a biased predictor to select hard examples. For all CV datasets and benchmarks, we extract 10\% data from the training set as a low-resource set for our main results. Each label has $500$ examples in CIFAR-10, $50$ examples in CIFAR-100, $100$ examples in ImageNet-1K.  

\paragraph{NLP}
We choose 8 datasets from GLUE~\citep{wang2018glue}, a collection understanding datasets for evaluating natural language understanding systems. Like CV, we select a subset of the full-size training set as an attack set while we set the number of data with the same labels as $k$. Following previous studies, we use the validation set as the test set considering the hidden test set. For the convenience of the demonstration, we show all the results with accuracy scores. For all NLP datasets, we implement BERT trained with one epoch as a biased predictor to select hard examples. For all NLP datasets, we extract 500 examples for each label (except for WNLI with 100 examples) as the training set for our main results. 

\subsection{Models}
We briefly describe the models in our benchmark. For all models, we run experiments three times and report the average results. Appendix~\ref{appendix: setup} provides detailed model descriptions and hyper-parameter settings. All models can achieve almost 100\% accuracy on the training sets. 
\paragraph{CV}
For over-parameterized neural networks, we consider the following models: 1) \textbf{FFN}, a feed-forward neural network; 2) \textbf{VGG-16}~\citep{vgg}, a classical convolutional neural network (CNN); 3) \textbf{ResNet-18}~\citep{DBLP:conf/cvpr/HeZRS16},  a residual neural network; 4) \textbf{DenseNet-121}~\citep{densenet}, a widely-used CNN architecture. Besides, we also re-implement two pre-trained models: 1) Transformer-based \textbf{ViT-B/16}~\citep{DBLP:conf/iclr/DosovitskiyB0WZ21} and 2) convolutional-based \textbf{EfficientNetV2-S}~\citep{DBLP:conf/icml/TanL21}.

\paragraph{NLP}
For NLP tasks, we implement the following models: 1) randomly-initialized \textbf{Transformer} encoder~\citep{vaswani2017attention}, the most popular backbone architecture; 2) \textbf{BERT} ~\citep{devlin2018bert}, a popular understanding model. 3)\textbf{GPT-2}~\citep{radford2019language}, a model based on the decoder of transformer. 4)\textbf{RoBERTa} ~\citep{liu2019roberta}, a robustly optimized BERT. 5)\textbf{T5} ~\citep{raffel2020exploring}, an encoder and decoder model.




\subsection{Results}

\paragraph{Hard-Bench challenges neural networks}
As Table~\ref{tab:02_main_nlp1}  and Table~\ref{tab:01_cv_main} illustrate, Hard-Bench can mislead neural networks with worse generalization errors. We re-implement strong understanding models, which have shown promising results in various low-resource tasks. For example, in Random-Bench, RoBERTa shows the near-human performance on SST2 with 91\% accuracy. However, the performance drops sharply on Hard-Bench with only 51.01\% accuracy on Hard-Bench (GradNorm) and 50.55\% accuracy on Hard-Bench (Loss), nearly random-guessing results.  Similar results are observed on CV datasets.   For example, DenseNet-121 trained on a random sampling set achieves high test results with $71.33\%$ accuracy on CIFAR-10. The accuracy drops to $59.87\%$ on Hard-Bench (GradNorm) and to $44.81\%$ on Hard-Bench (Loss). The large performance drop also indicates that there is still a large gap between existing models and human-level models. All these drops demonstrate that our benchmark poses a great challenge for neural networks. Due to space limitations, we report a default setting that selects fixed examples for each label. We also conduct more experiments with different shots. We find that ``hard examples'' especially challenge low-resource learning. The performance drop gradually becomes small with the increasing number of training data. More results can be found at Appendix D.  


\paragraph{Pre-trained networks show strong generation results on CV benchmarks, but still suffer from handling NLP tasks} Compared with randomly-initialized models, pre-trained networks show better generalization results in CV datasets, as shown in Table~\ref{tab:01_cv_main}. For example, EfficientNetV2-S does not yield obvious performance drops on Hard-Bench, with slight 2.63\% accuracy drop  on Hard-Bench (Loss) and 1.10\% accuracy gain on Hard-Bench (GradNorm). As a comparison, pre-trained networks have much worse results on NLP tasks.  On random-Bench, pre-trained networks bring large performance improvements over random-initialized baseline (Transformer). However, on our benchmark, all pre-trained networks yield surprising performance drops. They even do not beat randomly-initialized models. These results demonstrate that the results of pre-trained models on NLP tasks are more easily over-estimated.

\paragraph{Hard-Bench (Loss) is more challenging than Hard-Bench (GradNorm)}
We implement two metrics to select hard examples, including loss and gradient norm. Despite similar motivation, Hard-Bench (loss) is more challenging than Hard-Bench (GradNorm) according to our experimental results. On NLP tasks, Hard-Bench (loss) also witnesses the worst results. Loss is the most direct signal to see how neural networks understand an example. These difficult examples confuse neural networks, which barely learn core features. This learning weakness is not covered by  existing low-resource benchmarks. Hard-Bench provides a new perspective for understanding the learning abilities of different models. 

\paragraph{Hard-Bench brings new model rankings}  Table~\ref{tab:02_main_nlp1} shows that RoBERTa achieves the best average performance on Random-Bench. However, RoBERTa yields much worse results Hard-Bench (Loss), only better than BERT. T5 is a relatively robust model, with higher results among the 5 models. 

\paragraph{Data augmentation slightly improves results}  Table 3 shows the results with data augmentation techniques. We apply a widely-used data augmentation method, cutmix~\citep{yun2019cutmix}, on all CV models and report the results on CIFAR-10. We can see that data augmentation brings slight performance improvements, but also faces the challenges of generalization on our benchmarks.
\begin{table}[ht]
\centering
\caption{
    Results with cutmix. Models with data augmentation still face the challenges of generalization on our benchmarks.
}
\label{tab:dataaug1}
     \footnotesize
  \resizebox{\linewidth}{!}{%
		\begin{tabular}{l|ccc}
			\toprule
\textbf{Models} & \textbf{\methodRandom} & \textbf{\methodgd} & \textbf{\methodls} \\
\midrule

 FFN              &  53.99\std{0.39} &  30.36\std{1.26} &  19.29\std{0.36} \\
 VGG-16           &  66.76\std{0.59} &  47.85\std{0.97} &  33.64\std{0.25} \\
 ResNet-18        &  68.94\std{0.66} &  52.73\std{1.54} &  37.96\std{1.12} \\
 DenseNet-121     &  75.44\std{0.34} &  63.23\std{0.42} &  47.70\std{1.38} \\
 ViT-B/16         &  97.71\std{0.17} &  97.79\std{0.08} &  97.09\std{0.12} \\
 EfficientNetV2-S &  93.25\std{0.65} &  92.83\std{0.63} &  91.41\std{0.69} \\
			\bottomrule
		\end{tabular}
 }

\end{table}




\begin{figure}
    \centering
    \includegraphics[width=\linewidth]{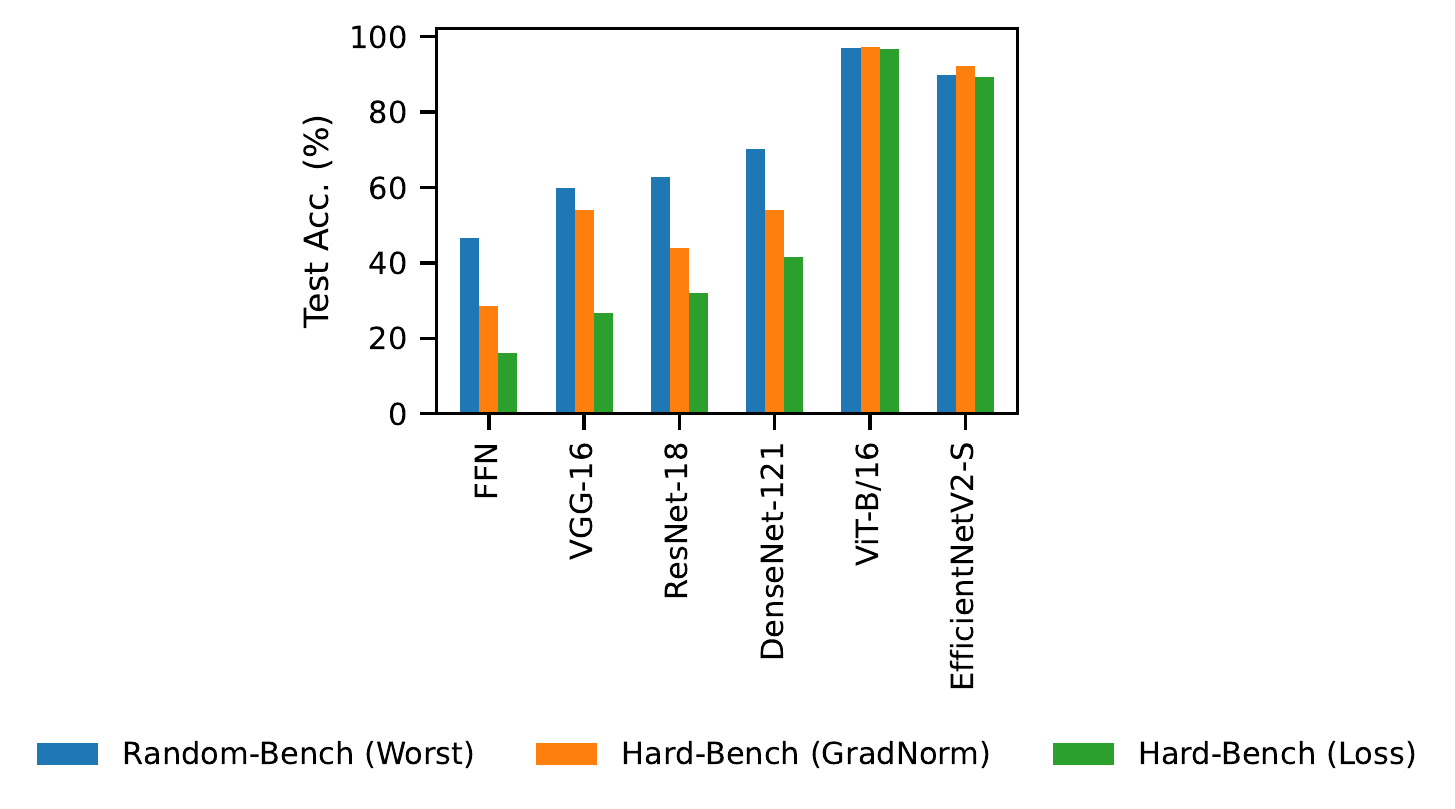}
    \caption{The worst performance on CIFAR 10. Massive sampling fails to find a challenging benchmark.}
    \label{fig:samp}
\end{figure}

\subsection{Ablation Studies}

\paragraph{Massive sampling fails to find a challenging benchmark}
In Random-Bench, we report the average results over 3 random samplings to show that random sampling is not insufficient to uncover the robust weakness of neural networks. In this part, we conduct 100 samplings and report the worst result in Figure~\ref{fig:samp} to figure out whether our methods can be replaced with massive sampling. As we can see, there is still a large gap between the worst results on Random-Bench and Hard-Bench, indicating that the proposed method is an effective method to build challenging benchmarks. 



\begin{table*}[t]
\centering
\caption{
    Results of Hard-Bench (Loss)  on CIFAR10 based on FFN, ResNet-18, and ViT-B/16 as predictors to select hard examples.  ``Gap'' represents the test accuracy gap with Bench-Random. Despite slight gap variance, all predictors bring large performance drops compared with Random-Bench. 
}
    \footnotesize
	 \resizebox{0.65\linewidth}{!}{
\begin{tabular}{l|cc|cc|cc}
			\toprule
 \multirow{2}{*}{\textbf{Models}}  & \multicolumn{2}{c|}{\textbf{FFN Predictor }} & \multicolumn{2}{c|}{\textbf{ResNet Predictor}} & \multicolumn{2}{c}{\textbf{ViT Predictor}} \\

   & \textbf{Accuracy} & \textbf{Gap} &  \textbf{Accuracy} & \textbf{Gap} &\textbf{Accuracy} & \textbf{Gap}  \\ 
			\midrule

FFN           &  29.64\std{  0.88} &\textbf{ \gap{ 19.37}} &  40.69\std{  0.69} & \gap{  8.32} &  39.82\std{  0.61} & \gap{  9.19}  \\
 VGG-16        &  55.11\std{  0.89} & \gap{ 13.52} &  51.83\std{  0.39} & \gap{ 16.80} &  48.19\std{  0.64} & \textbf{\gap{ 20.44}}  \\
 ResNet-18    &  46.87\std{  2.41} & \textbf{\gap{ 18.64}} &  53.93\std{  0.72} & \gap{ 11.58} &  50.59\std{  0.98} & \gap{ 14.92}  \\
 DenseNet-121  &  59.87\std{  0.66} & \gap{ 11.55} &  61.70\std{  0.23} & \gap{  9.72} &  58.05\std{  0.80} & \textbf{\gap{ 13.37}}  \\


 ViT-B/16            &  97.39\std{  0.10} & \gap{ -0.32} &  97.07\std{  0.19} & \gap{  0.00} &  96.92\std{  0.32} & \textbf{\gap{  0.15}}  \\
EfficientNet-V2    &  92.51\std{  0.25} & \gap{ -0.69} &  89.70\std{  0.32} & \gap{  2.12} &  87.26\std{  1.01} & \textbf{\gap{  4.56}}  \\

			\bottomrule
		\end{tabular}
	 }
	\vspace{-0.5ex}
	\label{tab:05_cv_diff}
\end{table*}

\begin{table*}[t]
\centering
\caption{
     Results of Hard-Bench (Loss) on CIFAR10 based on BERT, a randomly-initialized Transformer, and GPT2 as predictors to select hard examples.  ``Gap'' represents the test accuracy gap with Bench-Random. Despite slight gap variance, all predictors bring large performance drops compared with Random-Bench. 
}
    \footnotesize
	\resizebox{0.65\linewidth}{!}{
\begin{tabular}{l|cc|cc|cc}
			\toprule
 \multirow{2}{*}{\textbf{Models}}  & \multicolumn{2}{c|}{\textbf{BERT Predictor }} & \multicolumn{2}{c|}{\textbf{Transformer Predictor}} & \multicolumn{2}{c}{\textbf{GPT-2 Predictor}} \\

  & \textbf{Accuracy} & \textbf{Gap} &  \textbf{Accuracy} & \textbf{Gap} &\textbf{Accuracy} & \textbf{Gap}  \\ 
			\midrule
   
 Transformer  &   51.38\std{  0.40} & \gap{ 16.78} &  51.17\std{  0.17} & \gap{ 16.99} &  50.55\std{  0.73} & \textbf{\gap{ 17.61}}  \\
 BERT        &   45.64\std{  5.32} & \textbf{\gap{ 43.04}} &  51.06\std{  2.51} & \gap{ 37.62} &  48.30\std{  2.00} & \gap{ 40.38}  \\
 GPT-2        &   49.79\std{  2.06} & \gap{ 38.29} &  50.46\std{  3.20} & \gap{ 37.62} &  48.88\std{  4.00} & \textbf{\gap{ 39.20} } \\
 RoBERTa     &   50.55\std{  0.62} & \gap{ 40.99} &  54.72\std{  3.04} & \gap{ 36.82} &  48.33\std{  2.19} & \textbf{\gap{ 43.21}}  \\
 T5          &   49.86\std{  2.85} & \textbf{\gap{ 38.87}} &  60.48\std{  3.88} & \gap{ 28.25} &  56.03\std{  2.62} & \gap{ 32.70}  \\

			\bottomrule
		\end{tabular}
	 }
	\label{tab:05_nlp_diff}
\end{table*}

\paragraph{Ablation studies on different models as predictors} In our framework, we introduce a weak classifier as a biased predictor. For simplification, we choose FFN for CV datasets and BERT for NLP datasets. We conduct experiments on more networks to see whether the choice of predictors affects our conclusions. Table~\ref{tab:05_cv_diff} and Table~\ref{tab:05_nlp_diff} show the attack results on CIFAR10 and SST-2. For CV models, we test three models: FFN, ResNet-18 and ViT-B/16, as predictors. For SST-2, we test three models: BERT, randomly-initialized Transformer, and GPT2, as predictors. All models show consistent performance drops, indicating that our method is a universal model to generate challenging datasets to attack various models. 

\begin{figure}[t]
    \centering
    \includegraphics[width=\linewidth]{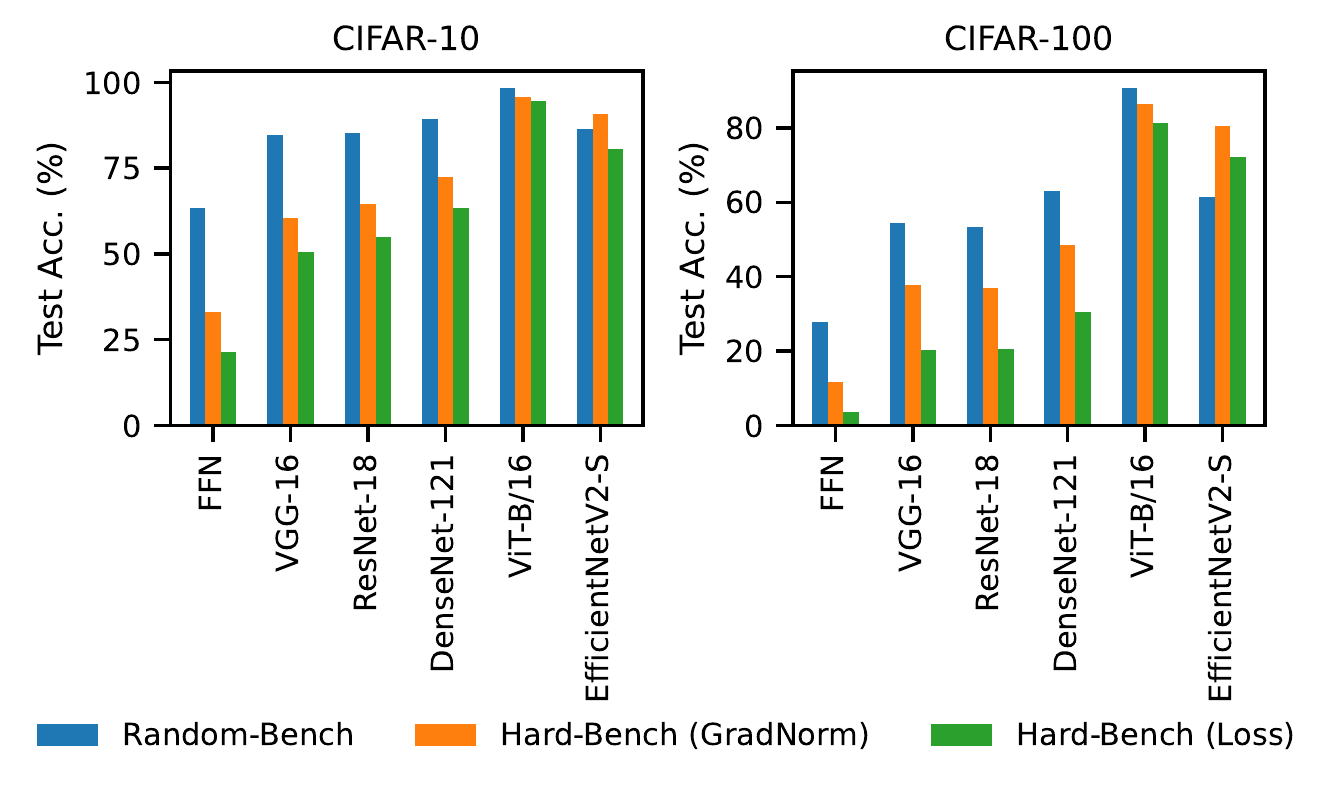}
    \caption{Results on the selected set as the test set. As we can see, these ``hard examples'' capture the weakness of neural networks. If neural networks do not see these results, they fail on handling these examples. }
    \label{fig:mylabel1}
\end{figure}

\paragraph{Results on the selected set as the test set} Figure~\ref{fig:mylabel1} shows results on the selected set as the test set. As we can
see, these ``hard examples'' capture the weakness of neural
networks. If neural networks do not see these results, they
fail on handling these examples.


\subsection{Explaining the Effectiveness of \BenchName with Visualization}
In this section, we compare samples selected by our \BenchName with samples from Random-Bench to demonstrate our approach reaches the goal of building difficult low-resource training set with shifted distributions. To make our observation more straightforward, we show visualizations in the Appendix \ref{visualizations}. We make the following observations based on these visualization results: 

\paragraph{\BenchName induces bias in the low-resource training set}
From visualizations, we can see that both GradNorm and Loss variations of \BenchName construct training sets that are drastically different from the data distribution. For the airplane class, the majority of data distribution shows the planes in a sky background, which is reflected in the random sampled sets. In contrast, there are no sky background plane figures in the samples selected by GradNorm, and most samples show a brown or green background. The Hard-Bench (Loss) on the other hand, prefers samples with white background, covering more than half of the selected set. Similar biases are evident in other classes, showing that our approach successfully induces bias in the low-resource training set.

\paragraph{\BenchName find challenging samples} Our method uses data difficulty metrics to find the most challenging cases in each dataset. This increase in data difficulty is distinct in both GradNorm and Loss variants. For the cat class, both GradNorm and Loss choose the figure of a cat tail, which is rare in the data distribution. Also, sketches of cats are more likely to be selected by our \BenchName than the Random-Bench. It shows that our proposed \BenchName has the potential to find in-domain yet challenging data samples. Similar patterns can be found in the samples for frogs (red frogs), cars (red and yellow beetles), and birds (kiwis).

\paragraph{It's harder to ``cheat'' Hard-Bench with spurious attributes} In this part, we stress the advantage of our benchmark to avoid the influence of spurious correlations. Spurious correlation occurs as a statistical phenomenon, whereas confounders in data can be used as shortcuts to perform inference. For example, as shown in Random-Bench Visualizations, 56\% of the airplane training set has a sky background, and the models easily overfit on this superficial feature, resulting in improved prediction figures with this feature in the test set (constitute the majority). However, Hard-Bench avoids spurious features by selecting less common figures.



\section{Conclusion}


This paper proposes a challenging benchmark for low-resource learning. We first analyze which factors affect the difficulty of low-resource learning. We prove that low-resource generalization results in worse performance with more difficult and biased datasets. So we choose two metrics for measuring data difficulty, which result in two variants, \methodls and \methodgd. Experiments show that both can better tell the learning gap between existing models than randomly-sampled low-resource datasets. We hope our work can encourage more low-resource robustness studies in the future work. 


\bibliography{icml2023}
\bibliographystyle{icml2023}

\newpage
\onecolumn
\appendix
\section{Perceptron Model of Low Resource Learning \label{appendix: perceptron theory}}

In this section, notations are defined as follows. We look at the teacher-student setting in learning perceptrons. Consider a large curated dataset of $N$ examples $D=\{x_i,y_i\}_{i\in [N]}$ where $x_i\in \mathbbm{R}^d$ are i.i.d. random Gaussian inputs $x_i \sim \mathcal{N}(0, I_d)$, with labels generated by a teacher perceptron $T\in \mathbbm{R}^d$ as $y_i = \text{sign}(Tx_i)$. The number of samples $N\rightarrow \infty$ but sample per parameter $\alpha = \frac{N}{d}=O(1)$ to remain trainable. Now we consider the low resource scenario where the number of training samples available $P$ is much less than $N$, where $\alpha_{\text{low}}=\frac{P}{d}\rightarrow 0$. For convenience, we sample the data for low resource learning from dataset $D$ such that $D_{\text{low}}=\{x_{\mu},y_{\mu}\}_{i\in[P]}\subset D$. Learning on $D_{\text{low}}$, we obtain a new student perceptron $J$ that has generalization error $\epsilon_g$. 

In the basic low-resource learning scenario, we use a uniform sampling strategy to obtain $D_{\text{low}}$ from $D$. We model $\epsilon_g$ as a function of $\alpha_{\text{low}}$. The results are as follows.

\begin{lemma} \label{low-resource}\textit{(Low-resource Learning, \citet{seung1992statistical})} For student perceptron $J$ learned on high dimension dataset $D_{\text{low}}$, the generalization error satisfies, 
\begin{equation}
    \epsilon_g \propto \alpha_{\text{low}}^{-1}
\end{equation}
\end{lemma}

For other settings, the generalization error is only related to the angle between teacher model $T$ and learned student model $J$. $\epsilon_g =\arccos{R}/ \pi$, $R = \frac{JT}{|J||T|}$. Based on different low-resource dataset sampling strategies, we calculate the teacher-student overlap $R$ with the geometry of each dataset distribution. \citet{sorscher2022beyond} has proved similar results in data pruning with our hard and biased learning setting. Here we only cite their results and don't elaborate on proofs. Note that despite the proofs being similar, we use a different setting in perceptron learning.  Their main objective is to understand how data-pruning can improve data efficiency, while we take an inverse stand, trying to understand challenging settings of low-resource learning. 

\begin{lemma} \label{low-resource}\textit{(Hard Low-resource Learning, \citet{sorscher2022beyond})} $D_{\text{low}}$ is sampled from $D$ such that $\forall{x_\mu \in D_{\text{low}}}, \forall{x_\gamma \in D/D_{\text{low}}}$, their margins satisfy $|Tx_\mu| \geq |Tx_\mu|$. Let $J$ be the student perceptron learned on high dimension dataset $D_{\text{low}}$, and $\kappa$ be the minimum margin $\min_{\mu}J(x^\mu y^\mu)$. If the perceptron is trained to maximum margin, the generalization error of $J$ satisfies, 
\begin{equation}
    \epsilon_g = \arccos{R}/\pi
\end{equation}
where R satisfies the saddle point equation,

\begin{equation}
\begin{aligned}
R=\frac{2\alpha}{f\sqrt{2\pi}\sqrt{1-R^{2}}}\int_{-\infty}^{\kappa}Dt\ \exp\bigg(-\frac{R^{2}t^{2}}{2(1-R^{2})}\bigg)\\ \cdot\bigg[1-\exp\bigg(-\frac{\gamma(\gamma-2Rt)}{2(1-R^{2})}\bigg)\bigg](\kappa-t) 
\end{aligned}
\end{equation}
in which $\gamma=H^{-1}(\frac{N-P}{2N})$, $p(z)=\frac{e^{-\frac{z^2}{2}}N}{\sqrt{2\pi} P}\Theta (\gamma -|z|)$
\end{lemma}

The proof for this lemma can be found in \citet{sorscher2022beyond} A.5.1 and is omitted here for brevity. 


\section{Low-Resource Generalization}

\subsection{Proof for Theorem ~\ref{covariate shift}\label{section: proof theorem}}

\begin{lemma}\label{mmd div} Define $\epsilon_S(h,f):= E_{x\sim S}|\delta(h(x))-\delta(f(x))|$. For any hypothesis $h, h' \in \mathcal{H}$, there exists $\epsilon_H > 0$ which satisfies,

\begin{equation}
    |\epsilon_{P_{low}}(h,h') - \epsilon_P(h,h')| \leq  \text{MMD}(\mathcal{H}, P_{low}, P) + \frac{\epsilon_{\mathcal{H}}}{2}
\end{equation}
 $\epsilon_H$ is a constant for the complexity of hypothesis space.

\end{lemma}

\begin{lemma}\label{subset}
    Let $f_{low}$ be the trained classifier on the low resource distribution $P_{low}$, and $f$ be the trained classifier on distribution $P$. Since $P_{low}$ is formed by a subset of the training examples, when training error $\epsilon_{P}(f)\rightarrow 0 $ and  $\epsilon_{P_{low}}(f_{low})\rightarrow 0$, $\epsilon_{P_{low}}(f_{low}, f) \leq \epsilon_{\alpha}$, where $\epsilon_{\alpha}$ is a constant approaching zero.
\end{lemma}

\textit{Proof.}
\begin{equation}
    \begin{aligned}
        \left|\epsilon_{P_{I_k}}\left(h, h^{\prime}\right)-\epsilon_P\left(h, h^{\prime}\right)\right| & \leq \sup _{h, h^{\prime} \in \mathcal{H}}\left|\epsilon_{P_{I_k}}\left(h, h^{\prime}\right)-\epsilon_P\left(h, h^{\prime}\right)\right| \\
        &= \sup _{h, h^{\prime} \in \mathcal{H}}\left|\mathbf{P}_{\boldsymbol{x} \sim P_{I_k}}\left[\delta(h(\boldsymbol{x})) \neq \delta(h^{\prime}(\boldsymbol{x}))\right]-\mathbf{P}_{\boldsymbol{x} \sim P}\left[\delta(h(\boldsymbol{x})) \neq \delta(h^{\prime}(\boldsymbol{x}))\right]\right| \\
        &= \sup _{h, h^{\prime} \in \mathcal{H}}\left|\mathbf{P}_{\boldsymbol{x} \sim P_{I_k}}\left[h(\boldsymbol{x}) \neq h^{\prime}(\boldsymbol{x})\right]-\mathbf{P}_{\boldsymbol{x} \sim P}\left[h(\boldsymbol{x}) \neq h^{\prime}(\boldsymbol{x})\right]\right| \\
        &= \sup _{h, h^{\prime} \in \mathcal{H}}\left|\int_{\mathcal{X}} \mathbf{1}_{h(\boldsymbol{x}) \neq h^{\prime}(\boldsymbol{x})} d \mu_{P_{I_k}}-\int_{\mathcal{X}} \mathbf{1}_{h(\boldsymbol{x}) \neq h^{\prime}(\boldsymbol{x})} d \mu_P\right|
    \end{aligned}
\end{equation}

\begin{lemma}\label{margin}
    Let $f$ be the trained classifier on dataset $D$ that is drawn i.i.d. from distribution $P$. $f$ is tested on a test dataset $S$ that is also drawn i.i.d. from the distribution $P$. Let $m$ be the maximum margin of classifier $f$. Then with probability at least $1-\delta$, 
    \begin{equation}
        \epsilon_S(f) \leq \epsilon_D(f) + c \sqrt{\frac{|\mathcal{H}|\ln m+\ln \left(\frac{1}{\delta}\right)}{m}}
    \end{equation}
    where $\epsilon_D(f)$ is the error on training set, and  $\epsilon_S(f)$ be the error on test set.
\end{lemma}

Following \citet{ben2010theory}, we use Lemma \ref{mmd div} and \ref{subset} to prove Theorem \ref{covariate shift}.

\textit{Proof}

\begin{equation}
    \begin{aligned}
    \epsilon_Q(f_{low}) &\leq \epsilon_Q(f) + \epsilon_Q(f_{low}, f) \\
    &= \epsilon_Q(f) + \epsilon_{P_{low}}(f_{low}, f) + (\epsilon_{Q}(f_{low}, f) - \epsilon_{P}(f_{low}, f)) + (\epsilon_{P}(f_{low}, f) - \epsilon_{P_{low}}(f_{low}, f)) \\
    &\leq \epsilon_Q(f) + \epsilon_{P_{low}}(f_{low}, f) + |\epsilon_{P}(f_{low}, f) - \epsilon_{Q}(f_{low}, f)| + |\epsilon_{P_{low}}(f_{low}, f) - \epsilon_{P}(f_{low}, f)| \\
    & \leq \epsilon_Q(f) + \epsilon_{\alpha} + |\epsilon_{P}(f_{low}, f) - \epsilon_{Q}(f_{low}, f)| + \text{MMD}(P_{low}, P) + \epsilon_{\mathcal{H}}
    \end{aligned}
\end{equation}

In which, 

\begin{equation}
    \begin{aligned}
    |\epsilon_{P}(f_{low}, f) - \epsilon_{Q}(f_{low}, f)| &= \left|\int_{\mathcal{X}} \mathbf{1}_{f_{low}(\boldsymbol{x}) \neq f(\boldsymbol{x})} d \mu_P-\int_{\mathcal{X}} \mathbf{1}_{f_{low}(\boldsymbol{x}) \neq f(\boldsymbol{x})} d \mu_Q\right| \\
    &= |\sum_{i=1}^n \mathbf{1}_{f_{low}(\boldsymbol{x_i}) \neq f(\boldsymbol{x_i})} - E_Q \mathbf{1}_{f_{low}(\boldsymbol{x}) \neq f(\boldsymbol{x})}|
    \end{aligned}
\end{equation}

Here suppose test set $Q$ matches the distribution of data for this classification task, and $P$ is constructed by sampling $n$ i.i.d. samples from the distribution $Q$. Using Lemma \ref{margin} we have, 
\begin{equation}
    P( |\epsilon_{P}(f_{low}, f) - \epsilon_{Q}(f_{low}, f)| > c\sqrt{\frac{|\mathcal{H}|\ln m+\ln \left(\frac{2}{\delta}\right)}{m}}) \leq \delta
\end{equation}

Therefore, with a probability over $1 - \delta$, we have
\begin{equation}
    \epsilon_Q(f_{low}) \leq \epsilon_Q(f) + \text{MMD}(P_{low}, P) + \epsilon_{\alpha} + \epsilon_{\mathcal{H}} + c\sqrt{\frac{|\mathcal{H}|\ln m+\ln \left(\frac{2}{\delta}\right)}{m}}
\end{equation}

\qedsymbol

\section{Models and Hyperparameters\label{appendix: setup}}

To evaluate the robustness of over-parameterized neural networks, we consider the following models.
1) \textbf{FFN}, a feed-forward neural network with two convolution and pooling layers and three feed-forward layers.
2) \textbf{VGG}~\citep{vgg}, a classical convolutional neural network. We use the VGG-16 with 13 convolution layers and three fully connected layers as implementation.
3) \textbf{ResNet}~\citep{DBLP:conf/cvpr/HeZRS16},  a residual neural network. We use the ResNet-18 with 16 residual blocks, one convolution layer, and one fully connected layer as implementation.
4)  \textbf{DenseNet}~\citep{densenet}. We use DenseNet-121 with 121 layers, one convolution layer, and one fully connected layer as re-implementation.\footnote{For VGG, ResNet, ResNeXt, and DenseNet on CIFAR and MNIST, we use the implementation from \url{https://github.com/kuangliu/pytorch-cifar}. As for ImageNet, we use the implementation from torch.models.}
Besides, to verify the attack ability \method  on the pre-trained models, we also re-implement two pre-trained models:
1) Transformer-based \textbf{ViT}~\citep{DBLP:conf/iclr/DosovitskiyB0WZ21}\footnote{We use the implementation from \url{https://huggingface.co/google/vit-base-patch16-224}} and
2) Convolutional-based \textbf{EfficientNetV2}~\citep{DBLP:conf/icml/TanL21}\footnote{We use the implementation from torch.models.}. 
For FFN, VGG, ResNet, ResNeXt, and DenseNet on ImageNet, we resize all the images into $256\times 256$ and then center-crop them into $224 \times 224$. For ViT on CIFAR, we resize all the images into $224 \times 224$, while $384\times 384$ for EfficientNetV2. 

We list hyper-paramter settings in Table~\ref{table:settings}.
All the SGD optimizers are with a momentum of $0.9$. For Adam/AdamW, we set $\beta = (0.9, 0.999)$. 
For the learning rate in selected Imagenet, the milestones are [10, 20].
We conduct all the experiments on a single A100 GPU.

We use Adam as the optimizer for all the NLP tasks with $learning\ rate = 2e-5$ and linear scheduler. We set all the batch sizes as $32$ and fine-tuned after the test accuracy no longer increased.

\begin{table*}[t]
\caption{
Hyperparameters settings. These settings are reported in their official repository for \emph{best practice}. The Linear refers LinearLR scheduler in Pytorch. OneCycle refers 1-cycle learning rate policy~\citep{OneCyclyLR}. 
}
\centering
\begin{tabular}{clcccc}
\toprule
\bf Models & \bf Datasets & \bf Batch Size & \bf Epochs & \bf Optimizer & \bf  Learning Rate \\
\midrule
\multirow{2}{*}{FFN}
~ & CIFAR-10 & 128 & 50 & Adam & [1e-3, 5e-4, 2.5e-4] \\
~ & CIFAR-100 & 128 & 50 & Adam & [1e-3, 5e-4, 2.5e-4]  \\
~ &  ImageNet-1K & 32 & 30 & SGD & [0.01, 0.001, 0.0001] \\
\midrule
\multirow{2}{*}{VGG}
~ & CIFAR-10 & 128 & 50 & Adam & [1e-4, 5e-5, 2.5e-5] \\
~ & CIFAR-100 & 128 & 50 & Adam & [1e-4, 5e-5, 2.5e-5]  \\
~ &  ImageNet-1K & 32 & 30 & SGD & [0.01, 0.001, 0.0001] \\
\midrule
\multirow{2}{*}{ResNet} 
~ & CIFAR-10 & 128 & 50 & Adam & [1e-3, 5e-4, 2.5e-4] \\
~ & CIFAR-100 & 128 & 50 & Adam & [1e-3, 5e-4, 2.5e-4]  \\
~ &  ImageNet-1K & 32 & 30 & SGD & [0.1, 0.01, 0.001] \\
\midrule
\multirow{2}{*}{DenseNet} 
~ & CIFAR-10 & 128 & 50 & Adam & [1e-3, 5e-4, 2.5e-4] \\
~ & CIFAR-100 & 128 & 50 & Adam & [1e-3, 5e-4, 2.5e-4]  \\
~ &  ImageNet-1K & 32 & 30 & SGD & [0.1, 0.01, 0.001] \\
\midrule
\multirow{2}{*}{ViT-B/16} & CIFAR-10 & 32  & 10 & Adam & $5\text{e-}5$ (Linear) \\
~ & CIFAR-100 & 32 &  10 &  Adam & $5\text{e-}5$ (Linear)  \\
\midrule
\multirow{2}{*}{EfficientNetV2-S} & CIFAR-10 & 32  & 10 & AdamW & $1\text{e-}3$ (OneCycle) \\
~ & CIFAR-100 & 32 &  10 &  AdamW & $1\text{e-}3$ (OneCycle)  \\
\bottomrule
\end{tabular}
\label{table:settings}

\end{table*}

\clearpage

\section{Results with Different Shots }
\label{appendix:multi_shots}

To better explore the effectiveness of \methodgd and \methodls, we demonstrate the results with different shots. The results are shown in the following tables.

\begin{table*}[htbp!]
\centering
\caption{
     Results on CIFAR-10. 
}
     \footnotesize
		\begin{tabular}{ll|c|cc|cc}
			\toprule
\multirow{2}{*}{\textbf{shots}}   & \multirow{2}{*}{\textbf{Models}} & \multirow{1}{*}{\textbf{\methodRandom}} & \multicolumn{2}{c|}{\textbf{\methodgd }} & \multicolumn{2}{c}{\textbf{\methodls}} \\

   & & \textbf{Accuracy} & \textbf{Accuracy} & \textbf{Gap} & \textbf{Accuracy} & \textbf{Gap} \\
			\midrule

\multirow{6}{*}{20-shot}  

& FFN               &  27.28\std{  1.51} &   13.68\std{  0.57} & \gap{ 13.60} &  10.74\std{  1.38} & \gap{ 16.54}  \\
& VGG-16            &  31.73\std{  1.26} &   14.76\std{  0.86} & \gap{ 16.97} &  10.27\std{  0.32} & \gap{ 21.46}  \\
& ResNet-18         &  30.54\std{  1.82} &   14.80\std{  0.56} & \gap{ 15.74} &  10.79\std{  0.35} & \gap{ 19.75}  \\
& DenseNet-121      &  34.69\std{  1.51} &   15.25\std{  0.29} & \gap{ 19.44} &  10.15\std{  0.61} & \gap{ 24.54}  \\


& ViT-B/16          &  79.84\std{  1.70} &   62.92\std{  1.97} & \gap{ 16.92} &  57.62\std{  2.46} & \gap{ 22.22}  \\
& EfficientNetV2-S  &  61.59\std{  4.36} &   40.67\std{  3.38} & \gap{ 20.92} &  31.44\std{  3.86} & \gap{ 30.15}  \\

			\midrule

\multirow{6}{*}{50-shot}

& FFN               &  33.31\std{  1.01} &   14.15\std{  0.71} & \gap{ 19.16} &   9.94\std{  0.98} & \gap{ 23.37}  \\
& VGG-16            &  38.95\std{  0.61} &   17.47\std{  0.96} & \gap{ 21.48} &  10.36\std{  0.45} & \gap{ 28.59}  \\
& ResNet-18         &  39.18\std{  1.19} &   17.78\std{  0.62} & \gap{ 21.40} &  10.64\std{  0.64} & \gap{ 28.54}  \\
& DenseNet-121      &  43.64\std{  0.68} &   18.56\std{  0.43} & \gap{ 25.08} &  10.15\std{  0.79} & \gap{ 33.49}  \\


& ViT-B/16          &  87.92\std{  0.45} &   82.77\std{  1.76} & \gap{  5.15} &  81.05\std{  2.08} & \gap{  6.87}  \\
& EfficientNetV2-S  &  74.75\std{  1.05} &   59.92\std{  3.84} & \gap{ 14.83} &  56.17\std{  3.56} & \gap{ 18.58}  \\

			\midrule

\multirow{6}{*}{200-shot}

& FFN               &  41.98\std{  0.79} &   21.05\std{  0.28} & \gap{ 20.93} &  13.11\std{  0.73} & \gap{ 28.87}  \\
& VGG-16            &  52.87\std{  0.78} &   25.29\std{  0.46} & \gap{ 27.58} &  15.35\std{  0.91} & \gap{ 37.52}  \\
& ResNet-18         &  53.67\std{  0.95} &   25.58\std{  0.42} & \gap{ 28.09} &  15.87\std{  0.51} & \gap{ 37.80}  \\
& DenseNet-121      &  61.69\std{  0.36} &   33.06\std{  1.57} & \gap{ 28.63} &  19.48\std{  0.86} & \gap{ 42.21}  \\


& ViT-B/16          &  95.30\std{  0.14} &   95.77\std{  0.19} & \gap{ -0.47} &  95.22\std{  0.26} & \gap{  0.08}  \\
& EfficientNetV2-S  &  88.25\std{  0.23} &   83.61\std{  1.24} & \gap{  4.64} &  82.28\std{  1.95} & \gap{  5.97}  \\

			\midrule

\multirow{6}{*}{2000-shot}

& FFN               &  58.83\std{  1.44} &   46.19\std{  0.66} & \gap{ 12.64} &  44.56\std{  1.86} & \gap{ 14.27}  \\
& VGG-16            &  78.50\std{  0.59} &   77.58\std{  0.40} & \gap{  0.92} &  76.34\std{  0.55} & \gap{  2.16}  \\
& ResNet-18         &  79.40\std{  0.35} &   79.00\std{  0.37} & \gap{  0.40} &  78.14\std{  0.17} & \gap{  1.26}  \\
& DenseNet-121      &  84.65\std{  0.34} &   84.70\std{  0.17} & \gap{ -0.05} &  83.58\std{  0.30} & \gap{  1.07}  \\


& ViT-B/16          &  97.86\std{  0.09} &   98.06\std{  0.12} & \gap{ -0.20} &  98.01\std{  0.08} & \gap{ -0.15}  \\
& EfficientNetV2-S  &  95.30\std{  0.18} &   95.79\std{  0.09} & \gap{ -0.49} &  95.07\std{  0.16} & \gap{  0.23}  \\

			\bottomrule
		\end{tabular}
	\vspace{-0.5ex}
	\label{tab:04_cifar10}
\end{table*}

\begin{table*}[htbp!]
\centering
\caption{
     Resuls on CIFAR100. 
}
     \footnotesize
		\begin{tabular}{ll|c|cc|cc}
			\toprule
\multirow{2}{*}{\textbf{shots}}   & \multirow{2}{*}{\textbf{Models}} & \multirow{1}{*}{\textbf{\methodRandom}} & \multicolumn{2}{c|}{\textbf{\methodgd }} & \multicolumn{2}{c}{\textbf{\methodls}} \\

   & & \textbf{Accuracy} & \textbf{Accuracy} & \textbf{Gap} & \textbf{Accuracy} & \textbf{Gap} \\
			\midrule

\multirow{6}{*}{20-shot}  

& FFN               &  10.48\std{  0.32} &    5.69\std{  0.11} & \gap{  4.79} &   1.90\std{  0.20} & \gap{  8.58}  \\
& VGG-16            &  18.87\std{  0.46} &   10.42\std{  0.17} & \gap{  8.45} &   2.77\std{  0.16} & \gap{ 16.10}  \\
& ResNet-18         &  17.20\std{  0.57} &    9.01\std{  0.34} & \gap{  8.19} &   2.61\std{  0.11} & \gap{ 14.59}  \\
& DenseNet-121      &  21.48\std{  0.80} &   10.84\std{  0.88} & \gap{ 10.64} &   3.24\std{  0.23} & \gap{ 18.24}  \\


& ViT-B/16          &  68.23\std{  1.68} &   61.45\std{  4.99} & \gap{  6.78} &  54.99\std{  2.04} & \gap{ 13.24}  \\
& EfficientNetV2-S  &  19.16\std{  0.58} &   51.87\std{  1.43} & \gap{-32.71} &  40.68\std{  1.30} & \gap{-21.52}  \\

\midrule

\multirow{6}{*}{200-shot}

& FFN               &  23.67\std{  1.00} &   16.83\std{  0.46} & \gap{  6.84} &  12.91\std{  1.45} & \gap{ 10.76}  \\
& VGG-16            &  45.52\std{  0.45} &   41.41\std{  0.77} & \gap{  4.11} &  36.22\std{  0.34} & \gap{  9.30}  \\
& ResNet-18         &  44.37\std{  0.70} &   41.03\std{  1.03} & \gap{  3.34} &  36.95\std{  1.02} & \gap{  7.42}  \\
& DenseNet-121      &  53.75\std{  0.39} &   51.80\std{  0.61} & \gap{  1.95} &  48.04\std{  0.37} & \gap{  5.71}  \\


& ViT-B/16          &  88.89\std{  0.24} &   89.00\std{  0.21} & \gap{ -0.11} &  88.86\std{  0.40} & \gap{  0.03}  \\
& EfficientNetV2-S  &  79.63\std{  0.64} &   80.36\std{  0.45} & \gap{ -0.73} &  78.32\std{  0.43} & \gap{  1.31}  \\

			\bottomrule
		\end{tabular}
	\vspace{-0.5ex}
	\label{tab:04_cifar100}
\end{table*}

\begin{table*}[htbp!]
\centering
\caption{
    Results on ImageNet. 
}
     \footnotesize
		\begin{tabular}{ll|c|cc|cc}
			\toprule
\multirow{2}{*}{\textbf{shots}}   & \multirow{2}{*}{\textbf{Models}} & \multirow{1}{*}{\textbf{\methodRandom}} & \multicolumn{2}{c|}{\textbf{\methodgd }} & \multicolumn{2}{c}{\textbf{\methodls}} \\

   & & \textbf{Accuracy} & \textbf{Accuracy} & \textbf{Gap} & \textbf{Accuracy} & \textbf{Gap} \\
			\midrule
      
\multirow{4}{*}{50-shot}  

& FFN               &   3.93\std{  0.51} &    1.59\std{  0.09} & \gap{  2.34} &   1.72\std{  0.04} & \gap{  2.21}  \\
& VGG-16            &   6.94\std{  0.43} &    2.11\std{  0.23} & \gap{  4.83} &   2.97\std{  0.23} & \gap{  3.97}  \\
& ResNet-18         &  18.84\std{  0.46} &   11.67\std{  0.27} & \gap{  7.17} &   9.46\std{  0.24} & \gap{  9.38}  \\
& DenseNet-121      &  22.96\std{  0.44} &   13.89\std{  0.45} & \gap{  9.07} &  10.29\std{  0.20} & \gap{ 12.67}  \\

			\bottomrule
		\end{tabular}
	\vspace{-0.5ex}
	\label{tab:04_imagenet}
\end{table*} 

 \begin{table*}[htbp!]
	\centering
	\caption{
		The results on GLUE with 16-shots. For more comprehensive comparison, we also report the results on two large language models, FLAN-T5 (11B) and BLOOM (1B). By leveraging instruction-tuning on the GLUE tasks, FLAN-T5 consistently achieves outstanding performance, suggesting that it is highly effective at completing the corresponding task. As a result, any attempt to attack the model is likely to fail.
	}
    \footnotesize
		\begin{tabular}{ll|c|cc|cc}
			\toprule
\multirow{2}{*}{\textbf{Datasets}}   & \multirow{2}{*}{\textbf{Models}} & \multirow{1}{*}{\textbf{\methodRandom}} & \multicolumn{2}{c|}{\textbf{\methodgd }} & \multicolumn{2}{c}{\textbf{\methodls}} \\

   & & \textbf{Accuracy} & \textbf{Accuracy} & \textbf{Gap} & \textbf{Accuracy} & \textbf{Gap} \\
			\midrule
           
   \multirow{5}{*}{SST2}
& Transformer &  52.64\std{  2.16} &   52.89\std{  0.56} & \gap{ -0.25} &  52.34\std{  0.26} & \gap{  0.30}  \\
& BERT        &  68.39\std{  7.14} &   56.86\std{  4.88} & \gap{ 11.53} &  50.28\std{  0.88} & \gap{ 18.11}  \\
& GPT-2        &  55.62\std{  4.12} &   52.52\std{  2.00} & \gap{  3.10} &  51.54\std{  1.22} & \gap{  4.08}  \\
& RoBERTa     &  76.67\std{  3.44} &   58.12\std{  1.47} & \gap{ 18.55} &  50.25\std{  0.96} & \gap{ 26.42}  \\
& T5          &  55.94\std{  3.74} &   51.95\std{  1.90} & \gap{  3.99} &  51.19\std{  2.26} & \gap{  4.75}  \\
\cmidrule{2-7}
& BLOOM          & 50.46 &   51.15  & \gap{  -0.69} & 50.11 & \gap{ 0.35 }\\
\midrule

\multirow{5}{*}{COLA}
& Transformer &  68.95\std{  0.44} &   68.74\std{  0.86} & \gap{  0.21} &  68.88\std{  0.50} & \gap{  0.07}  \\
& BERT        &  66.94\std{  3.55} &   64.99\std{  6.39} & \gap{  1.95} &  58.16\std{ 13.44} & \gap{  8.78}  \\
& GPT-2        &  66.40\std{  5.50} &   66.19\std{  5.92} & \gap{  0.21} &  66.56\std{  5.19} & \gap{ -0.16}  \\
& RoBERTa     &  69.66\std{  1.02} &   65.23\std{  5.14} & \gap{  4.43} &  49.38\std{ 11.73} & \gap{ 20.28}  \\
& T5          &  55.82\std{  8.92} &   59.54\std{  4.84} & \gap{ -3.72} &  56.80\std{  6.79} & \gap{ -0.98}  \\
\cmidrule{2-7}
& BLOOM          &   60.40 &  33.34  & \gap{ 27.06 } &46.40  & \gap{  14.00}  \\
\midrule

\multirow{5}{*}{MNLI}
& Transformer &  35.40\std{  0.09} &   35.45\std{  0.00} & \gap{ -0.05} &  35.28\std{  0.21} & \gap{  0.12}  \\
& BERT        &  36.21\std{  0.96} &   34.21\std{  0.54} & \gap{  2.00} &  34.03\std{  0.96} & \gap{  2.18}  \\
& GPT-2        &  37.63\std{  1.29} &   34.40\std{  1.39} & \gap{  3.23} &  33.88\std{  1.31} & \gap{  3.75}  \\
& RoBERTa     &  43.13\std{  2.07} &   35.48\std{  0.82} & \gap{  7.65} &  33.38\std{  1.10} & \gap{  9.75}  \\
& T5          &  33.98\std{  0.50} &   33.44\std{  0.21} & \gap{  0.54} &  33.39\std{  0.18} & \gap{  0.59}  \\
\cmidrule{2-7}
& BLOOM          & 35.42  & 35.33 & \gap{ 0.09 }& 35.42 &\gap{ 0.00 }\\
\midrule


\multirow{5}{*}{QNLI}
& Transformer &  53.48\std{  2.46} &   50.95\std{  0.55} & \gap{  2.53} &  51.22\std{  0.38} & \gap{  2.26}  \\
& BERT        &  53.75\std{  0.69} &   50.79\std{  0.31} & \gap{  2.96} &  50.10\std{  0.66} & \gap{  3.65}  \\
& GPT-2        &  55.16\std{  3.26} &   53.65\std{  2.94} & \gap{  1.51} &  52.49\std{  2.08} & \gap{  2.67}  \\
& RoBERTa     &  63.52\std{  3.92} &   50.80\std{  0.37} & \gap{ 12.72} &  49.78\std{  0.40} & \gap{ 13.74}  \\
& T5          &  54.03\std{  2.36} &   50.85\std{  1.02} & \gap{  3.18} &  49.69\std{  0.94} & \gap{  4.34}  \\
\cmidrule{2-7}
& BLOOM          & 50.54  &  50.47 &\gap{ 0.07 } & 50.01&\gap{ 0.53 }\\
\midrule

\multirow{5}{*}{MRPC}
& Transformer &  68.63\std{  0.31} &   68.38\std{  0.00} & \gap{  0.25} &  68.33\std{  0.10} & \gap{  0.30}  \\
& BERT        &  66.47\std{  3.22} &   63.19\std{  6.52} & \gap{  3.28} &  54.61\std{ 16.68} & \gap{ 11.86}  \\
& GPT-2        &  67.75\std{  1.53} &   66.23\std{  4.44} & \gap{  1.52} &  63.87\std{  8.90} & \gap{  3.88}  \\
& RoBERTa     &  69.26\std{  1.48} &   57.60\std{  7.79} & \gap{ 11.66} &  33.33\std{  0.83} & \gap{ 35.93}  \\
& T5          &  58.58\std{  5.94} &   59.90\std{  4.06} & \gap{ -1.32} &  56.32\std{  8.15} & \gap{  2.26}  \\
\cmidrule{2-7}
& BLOOM          & 66.18  & 65.44  &\gap{ 0.74 }& 65.93 &\gap{ 0.25 }\\
\midrule

\multirow{5}{*}{QQP}
& Transformer &  63.75\std{  0.55} &   63.23\std{  0.08} & \gap{  0.52} &  63.19\std{  0.02} & \gap{  0.56}  \\
& BERT        &  64.81\std{  2.15} &   59.19\std{  2.86} & \gap{  5.62} &  57.27\std{  4.69} & \gap{  7.54}  \\
& GPT-2        &  62.57\std{  1.34} &   54.64\std{  4.89} & \gap{  7.93} &  56.84\std{  3.40} & \gap{  5.73}  \\
& RoBERTa     &  65.55\std{  1.36} &   63.18\std{  0.00} & \gap{  2.37} &  63.10\std{  0.10} & \gap{  2.45}  \\
& T5          &  55.49\std{  3.35} &   56.61\std{  3.44} & \gap{ -1.12} &  56.14\std{  2.58} & \gap{ -0.65}  \\
\cmidrule{2-7}
& BLOOM          & 51.79  & 60.78 &\gap{ -8.99 } &60.79& \gap{ -9.00 }\\
\midrule

\multirow{5}{*}{RTE}
& Transformer &  53.72\std{  0.90} &   54.95\std{  1.01} & \gap{ -1.23} &  54.80\std{  0.42} & \gap{ -1.08}  \\
& BERT        &  55.02\std{  1.56} &   53.43\std{  2.41} & \gap{  1.59} &  50.40\std{  2.59} & \gap{  4.62}  \\
& GPT-2        &  58.77\std{  3.98} &   58.84\std{  2.84} & \gap{ -0.07} &  52.71\std{  1.69} & \gap{  6.06}  \\
& RoBERTa     &  55.16\std{  1.73} &   53.14\std{  0.42} & \gap{  2.02} &  52.56\std{  0.37} & \gap{  2.60}  \\
& T5          &  51.05\std{  2.44} &   49.03\std{  1.76} & \gap{  2.02} &  49.10\std{  0.97} & \gap{  1.95}  \\
\cmidrule{2-7}
& BLOOM          & 52.70  & 47.29 & \gap{5.41  }  &47.29&\gap{ 5.41 } \\
\midrule

\multirow{5}{*}{WNLI}
& Transformer &  58.03\std{  2.07} &   56.62\std{  0.56} & \gap{  1.41} &  57.46\std{  1.05} & \gap{  0.57}  \\
& BERT        &  56.34\std{  6.96} &   54.37\std{  4.51} & \gap{  1.97} &  56.62\std{  2.87} & \gap{ -0.28}  \\
& GPT-2        &  56.34\std{  0.00} &   56.62\std{  1.38} & \gap{ -0.28} &  56.90\std{  2.61} & \gap{ -0.56}  \\
& RoBERTa     &  57.75\std{  3.09} &   56.90\std{  1.13} & \gap{  0.85} &  56.34\std{  1.54} & \gap{  1.41}  \\
& T5          &  58.31\std{  0.69} &   53.52\std{  5.12} & \gap{  4.79} &  52.11\std{  5.42} & \gap{  6.20}  \\
\cmidrule{2-7}
& BLOOM          & 42.25  &42.25  & \gap{  0.00} &43.66 & \gap{  -1.41}  \\

\bottomrule
		\end{tabular}%
	\vspace{-0.5ex}
	\label{tab:04_nlp_16}
\end{table*}

 \begin{table*}[htbp!]
	\centering
	\caption{
		 Results on GLUE with 32-shots. 
	}
    \footnotesize
		\begin{tabular}{ll|c|cc|cc}
			\toprule
\multirow{2}{*}{\textbf{Datasets}}   & \multirow{2}{*}{\textbf{Models}} & \multirow{1}{*}{\textbf{\methodRandom}} & \multicolumn{2}{c|}{\textbf{\methodgd }} & \multicolumn{2}{c}{\textbf{\methodls}} \\

   & & \textbf{Accuracy} & \textbf{Accuracy} & \textbf{Gap} & \textbf{Accuracy} & \textbf{Gap} \\
			\midrule
  
   \multirow{5}{*}{SST2}
& Transformer &  55.96\std{  1.33} &   52.50\std{  0.58} & \gap{  3.46} &  52.27\std{  0.48} & \gap{  3.69}  \\
& BERT        &  77.20\std{  4.97} &   54.70\std{  4.07} & \gap{ 22.50} &  50.28\std{  0.85} & \gap{ 26.92}  \\
& GPT-2        &  68.46\std{  5.20} &   57.73\std{  1.01} & \gap{ 10.73} &  51.72\std{  0.72} & \gap{ 16.74}  \\
& RoBERTa     &  83.81\std{  2.25} &   57.36\std{  2.45} & \gap{ 26.45} &  50.09\std{  0.97} & \gap{ 33.72}  \\
& T5          &  63.10\std{  3.97} &   54.06\std{  2.86} & \gap{  9.04} &  51.06\std{  2.38} & \gap{ 12.04}  \\

\midrule

\multirow{5}{*}{COLA}
& Transformer &  69.36\std{  0.37} &   69.13\std{  0.00} & \gap{  0.23} &  69.15\std{  0.04} & \gap{  0.21}  \\
& BERT        &  67.56\std{  3.19} &   62.05\std{  9.21} & \gap{  5.51} &  60.44\std{ 10.81} & \gap{  7.12}  \\
& GPT-2        &  66.94\std{  3.91} &   66.06\std{  5.77} & \gap{  0.88} &  66.04\std{  6.22} & \gap{  0.90}  \\
& RoBERTa     &  72.23\std{  1.32} &   66.27\std{  3.90} & \gap{  5.96} &  59.85\std{ 12.56} & \gap{ 12.38}  \\
& T5          &  59.50\std{  4.25} &   58.39\std{  5.32} & \gap{  1.11} &  57.81\std{  6.12} & \gap{  1.69}  \\

\midrule

\multirow{5}{*}{MNLI}
& Transformer &  35.46\std{  0.03} &   35.45\std{  0.00} & \gap{  0.01} &  35.42\std{  0.04} & \gap{  0.04}  \\
& BERT        &  39.21\std{  2.50} &   34.45\std{  0.76} & \gap{  4.76} &  33.41\std{  0.63} & \gap{  5.80}  \\
& GPT-2        &  39.81\std{  0.88} &   34.75\std{  1.47} & \gap{  5.06} &  34.00\std{  1.28} & \gap{  5.81}  \\
& RoBERTa     &  45.44\std{  2.02} &   36.55\std{  0.97} & \gap{  8.89} &  33.81\std{  1.18} & \gap{ 11.63}  \\
& T5          &  34.45\std{  0.50} &   33.87\std{  0.24} & \gap{  0.58} &  33.13\std{  0.24} & \gap{  1.32}  \\
\midrule


\multirow{5}{*}{QNLI}
& Transformer &  53.96\std{  0.92} &   51.36\std{  0.40} & \gap{  2.60} &  50.78\std{  0.09} & \gap{  3.18}  \\
& BERT        &  57.07\std{  2.02} &   50.85\std{  0.40} & \gap{  6.22} &  50.08\std{  0.35} & \gap{  6.99}  \\
& GPT-2        &  57.97\std{  2.83} &   53.70\std{  3.03} & \gap{  4.27} &  52.86\std{  2.50} & \gap{  5.11}  \\
& RoBERTa     &  71.64\std{  1.99} &   50.67\std{  0.62} & \gap{ 20.97} &  49.61\std{  0.31} & \gap{ 22.03}  \\
& T5          &  60.41\std{  4.20} &   50.74\std{  1.29} & \gap{  9.67} &  49.50\std{  1.04} & \gap{ 10.91}  \\
\midrule

\multirow{5}{*}{MRPC}
& Transformer &  68.63\std{  0.27} &   68.43\std{  0.10} & \gap{  0.20} &  68.48\std{  0.20} & \gap{  0.15}  \\
& BERT        &  66.96\std{  2.36} &   61.72\std{  7.05} & \gap{  5.24} &  54.80\std{ 16.04} & \gap{ 12.16}  \\
& GPT-2        &  69.07\std{  1.82} &   67.11\std{  2.33} & \gap{  1.96} &  63.97\std{  8.95} & \gap{  5.10}  \\
& RoBERTa     &  73.73\std{  2.79} &   55.34\std{  6.31} & \gap{ 18.39} &  35.39\std{  4.57} & \gap{ 38.34}  \\
& T5          &  62.21\std{  3.44} &   58.04\std{  5.53} & \gap{  4.17} &  55.78\std{  8.91} & \gap{  6.43}  \\

\midrule

\multirow{5}{*}{QQP}
& Transformer &  64.06\std{  0.30} &   63.26\std{  0.09} & \gap{  0.80} &  63.18\std{  0.00} & \gap{  0.88}  \\
& BERT        &  65.49\std{  1.73} &   61.12\std{  1.72} & \gap{  4.37} &  54.83\std{  5.47} & \gap{ 10.66}  \\
& GPT-2        &  63.37\std{  3.21} &   56.03\std{  4.17} & \gap{  7.34} &  55.00\std{  5.30} & \gap{  8.37}  \\
& RoBERTa     &  70.10\std{  0.98} &   61.82\std{  2.73} & \gap{  8.28} &  62.81\std{  0.45} & \gap{  7.29}  \\
& T5          &  61.44\std{  4.99} &   56.21\std{  2.96} & \gap{  5.23} &  54.53\std{  4.05} & \gap{  6.91}  \\
\midrule

\multirow{5}{*}{RTE}
& Transformer &  53.72\std{  0.98} &   55.38\std{  0.49} & \gap{ -1.66} &  55.02\std{  0.74} & \gap{ -1.30}  \\
& BERT        &  55.02\std{  3.83} &   52.85\std{  1.96} & \gap{  2.17} &  49.46\std{  2.45} & \gap{  5.56}  \\
& GPT-2        &  58.77\std{  2.65} &   60.36\std{  2.96} & \gap{ -1.59} &  52.85\std{  3.15} & \gap{  5.92}  \\
& RoBERTa     &  57.76\std{  3.62} &   54.95\std{  2.37} & \gap{  2.81} &  52.78\std{  0.58} & \gap{  4.98}  \\
& T5          &  51.48\std{  1.13} &   52.06\std{  1.77} & \gap{ -0.58} &  49.17\std{  1.60} & \gap{  2.31}  \\
\midrule

\multirow{5}{*}{WNLI}
& Transformer &  58.59\std{  2.76} &   56.34\std{  0.00} & \gap{  2.25} &  58.31\std{  1.44} & \gap{  0.28}  \\
& BERT        &  54.08\std{  3.94} &   54.93\std{  4.27} & \gap{ -0.85} &  55.21\std{  2.87} & \gap{ -1.13}  \\
& GPT-2        &  58.03\std{  2.25} &   57.46\std{  1.87} & \gap{  0.57} &  56.34\std{  1.99} & \gap{  1.69}  \\
& RoBERTa     &  56.62\std{  0.56} &   56.90\std{  1.13} & \gap{ -0.28} &  57.18\std{  1.44} & \gap{ -0.56}  \\
& T5          &  53.80\std{  3.92} &   57.18\std{  3.03} & \gap{ -3.38} &  53.24\std{  5.52} & \gap{  0.56}  \\

\bottomrule
		\end{tabular}%
	
	\vspace{-0.5ex}
	\label{tab:04_nlp_32}
\end{table*}

\begin{table*}[htbp!]
	\centering
	\caption{
		Results on GLUE with 100-shots. 
	}
    \footnotesize
		\begin{tabular}{ll|c|cc|cc}
			\toprule
\multirow{2}{*}{\textbf{Datasets}}   & \multirow{2}{*}{\textbf{Models}} & \multirow{1}{*}{\textbf{\methodRandom}} & \multicolumn{2}{c|}{\textbf{\methodgd }} & \multicolumn{2}{c}{\textbf{\methodls}} \\

   & & \textbf{Accuracy} & \textbf{Accuracy} & \textbf{Gap} & \textbf{Accuracy} & \textbf{Gap} \\
			\midrule
     
   \multirow{5}{*}{SST2}
& Transformer &  59.50\std{  1.52} &   52.41\std{  0.64} & \gap{  7.09} &  51.74\std{  0.38} & \gap{  7.76}  \\
& BERT        &  86.22\std{  0.39} &   51.38\std{  1.86} & \gap{ 34.84} &  49.33\std{  2.12} & \gap{ 36.89}  \\
& GPT-2        &  83.00\std{  1.43} &   53.46\std{  1.76} & \gap{ 29.54} &  51.22\std{  1.85} & \gap{ 31.78}  \\
& RoBERTa     &  88.37\std{  1.08} &   51.93\std{  0.73} & \gap{ 36.44} &  50.57\std{  0.75} & \gap{ 37.80}  \\
& T5          &  83.35\std{  4.21} &   52.25\std{  1.54} & \gap{ 31.10} &  51.01\std{  2.49} & \gap{ 32.34}  \\
\midrule

\multirow{5}{*}{COLA}
& Transformer &  69.19\std{  0.08} &   69.17\std{  0.05} & \gap{  0.02} &  68.78\std{  0.69} & \gap{  0.41}  \\
& BERT        &  74.84\std{  1.36} &   61.25\std{  7.11} & \gap{ 13.59} &  57.09\std{ 12.93} & \gap{ 17.75}  \\
& GPT-2        &  66.62\std{  3.15} &   65.77\std{  5.94} & \gap{  0.85} &  64.60\std{  6.87} & \gap{  2.02}  \\
& RoBERTa     &  77.28\std{  1.09} &   62.84\std{  6.71} & \gap{ 14.44} &  59.64\std{  5.95} & \gap{ 17.64}  \\
& T5          &  75.24\std{  1.07} &   57.49\std{  5.44} & \gap{ 17.75} &  56.72\std{  7.18} & \gap{ 18.52}  \\
\midrule

\multirow{5}{*}{MNLI}
& Transformer &  35.61\std{  0.22} &   35.25\std{  0.32} & \gap{  0.36} &  35.11\std{  0.41} & \gap{  0.50}  \\
& BERT        &  43.98\std{  2.71} &   34.74\std{  0.62} & \gap{  9.24} &  33.13\std{  0.60} & \gap{ 10.85}  \\
& GPT-2        &  48.69\std{  1.80} &   34.77\std{  1.12} & \gap{ 13.92} &  33.86\std{  1.24} & \gap{ 14.83}  \\
& RoBERTa     &  61.44\std{  2.69} &   36.87\std{  1.00} & \gap{ 24.57} &  33.64\std{  0.98} & \gap{ 27.80}  \\
& T5          &  40.63\std{  4.32} &   34.27\std{  0.36} & \gap{  6.36} &  32.98\std{  0.29} & \gap{  7.65}  \\
\midrule


\multirow{5}{*}{QNLI}
& Transformer &  56.23\std{  0.86} &   50.68\std{  0.15} & \gap{  5.55} &  50.54\std{  0.00} & \gap{  5.69}  \\
& BERT        &  63.82\std{  4.63} &   49.98\std{  1.01} & \gap{ 13.84} &  47.26\std{  2.17} & \gap{ 16.56}  \\
& GPT-2        &  62.52\std{  4.58} &   53.34\std{  2.93} & \gap{  9.18} &  51.80\std{  2.02} & \gap{ 10.72}  \\
& RoBERTa     &  78.44\std{  2.05} &   50.08\std{  0.51} & \gap{ 28.36} &  50.23\std{  0.46} & \gap{ 28.21}  \\
& T5          &  73.75\std{  2.43} &   49.99\std{  1.01} & \gap{ 23.76} &  48.91\std{  1.36} & \gap{ 24.84}  \\
\midrule

\multirow{5}{*}{MRPC}
& Transformer &  69.02\std{  0.69} &   68.43\std{  0.10} & \gap{  0.59} &  66.67\std{  3.43} & \gap{  2.35}  \\
& BERT        &  69.41\std{  0.95} &   58.77\std{  5.61} & \gap{ 10.64} &  48.24\std{ 11.18} & \gap{ 21.17}  \\
& GPT-2        &  71.76\std{  1.91} &   65.34\std{  2.14} & \gap{  6.42} &  64.07\std{  5.23} & \gap{  7.69}  \\
& RoBERTa     &  77.16\std{  2.77} &   60.34\std{  4.15} & \gap{ 16.82} &  41.47\std{  6.42} & \gap{ 35.69}  \\
& T5          &  65.64\std{  1.33} &   57.79\std{  5.58} & \gap{  7.85} &  56.52\std{  5.77} & \gap{  9.12}  \\
\midrule

\multirow{5}{*}{QQP}
& Transformer &  65.27\std{  0.61} &   63.45\std{  0.32} & \gap{  1.82} &  60.54\std{  1.67} & \gap{  4.73}  \\
& BERT        &  69.62\std{  1.85} &   60.11\std{  1.68} & \gap{  9.51} &  47.14\std{  5.08} & \gap{ 22.48}  \\
& GPT-2        &  70.13\std{  2.28} &   55.04\std{  4.57} & \gap{ 15.09} &  51.27\std{  6.85} & \gap{ 18.86}  \\
& RoBERTa     &  75.33\std{  1.24} &   63.02\std{  1.13} & \gap{ 12.31} &  48.10\std{ 12.34} & \gap{ 27.23}  \\
& T5          &  73.02\std{  1.75} &   57.54\std{  4.38} & \gap{ 15.48} &  53.35\std{  4.40} & \gap{ 19.67}  \\

\midrule

\multirow{5}{*}{RTE}
& Transformer &  53.72\std{  0.90} &   56.10\std{  0.71} & \gap{ -2.38} &  53.07\std{  0.23} & \gap{  0.65}  \\
& BERT        &  54.66\std{  2.65} &   52.56\std{  1.06} & \gap{  2.10} &  47.44\std{  1.49} & \gap{  7.22}  \\
& GPT-2        &  59.35\std{  3.23} &   57.76\std{  2.33} & \gap{  1.59} &  51.26\std{  2.40} & \gap{  8.09}  \\
& RoBERTa     &  63.39\std{  2.49} &   53.94\std{  0.67} & \gap{  9.45} &  51.05\std{  2.19} & \gap{ 12.34}  \\
& T5          &  53.72\std{  3.97} &   51.19\std{  1.59} & \gap{  2.53} &  48.59\std{  1.32} & \gap{  5.13}  \\

\bottomrule
		\end{tabular}%
	\vspace{-0.5ex}
	\label{tab:04_nlp_100}
\end{table*}

\clearpage

\section{Visualization \label{visualizations}}
\subsection{Visualization of the ``Average Examples (Random)''}

\begin{figure}[htbp!]
    \caption{Visualization of the set random selected on CIFAR-10. We sample randomly 50 examples for each label.}
    \begin{minipage}[b]{0.5\textwidth}
     \centering
        \includegraphics[scale=0.6]{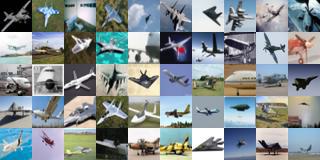}

    \end{minipage}
    \begin{minipage}[b]{0.5\textwidth}
     \centering
        \includegraphics[scale=0.6]{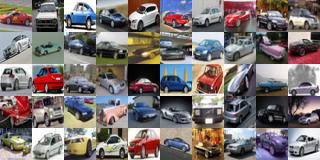}

    \end{minipage}
    
	\vspace{1.5ex}
    \begin{minipage}[b]{0.5\textwidth}
     \centering
        \includegraphics[scale=0.6]{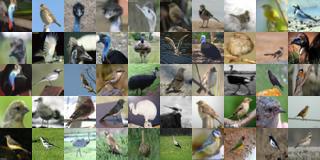}

    \end{minipage}
    \begin{minipage}[b]{0.5\textwidth}
     \centering
        \includegraphics[scale=0.6]{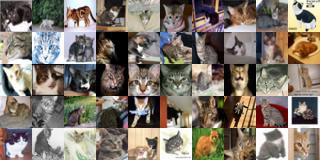}

    \end{minipage}

	\vspace{1.5ex}
    \begin{minipage}[b]{0.5\textwidth}
     \centering
        \includegraphics[scale=0.6]{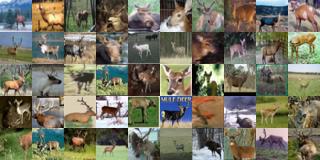}

    \end{minipage}
    \begin{minipage}[b]{0.5\textwidth}
     \centering
        \includegraphics[scale=0.6]{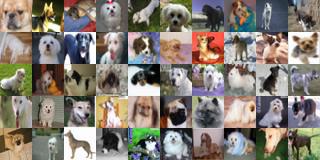}

    \end{minipage}
    
	\vspace{1.5ex}
    \begin{minipage}[b]{0.5\textwidth}
     \centering
        \includegraphics[scale=0.6]{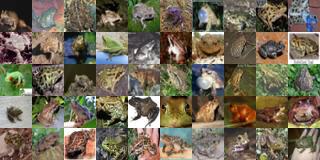}

    \end{minipage}
    \begin{minipage}[b]{0.5\textwidth}
     \centering
        \includegraphics[scale=0.6]{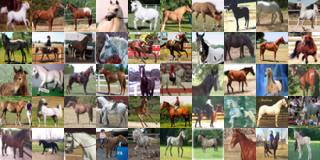}

    \end{minipage}
    
	\vspace{1.5ex}
    \begin{minipage}[b]{0.5\textwidth}
     \centering
        \includegraphics[scale=0.6]{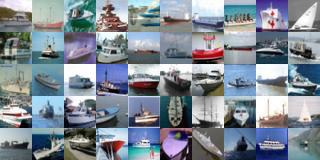}

    \end{minipage}
    \begin{minipage}[b]{0.5\textwidth}
     \centering
        \includegraphics[scale=0.6]{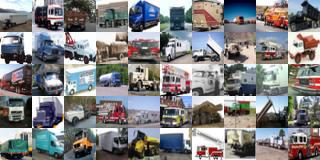}

    \end{minipage}
    \label{fig: imagenet-spurious2}
\end{figure}

\clearpage

\subsection{Visualization of the Searched ``Hard Examples (GradNorm)''}

\begin{figure}[htbp!]
    \caption{Visualization of the set searched by \methodgd on CIFAR-10. We choose top 50 examples for each label.}
    \begin{minipage}[b]{0.5\textwidth}
     \centering
        \includegraphics[scale=0.6]{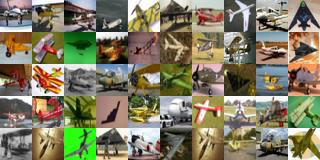}

    \end{minipage}
    \begin{minipage}[b]{0.5\textwidth}
     \centering
        \includegraphics[scale=0.6]{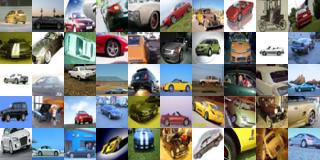}

    \end{minipage}
    
	\vspace{1.5ex}
    \begin{minipage}[b]{0.5\textwidth}
     \centering
        \includegraphics[scale=0.6]{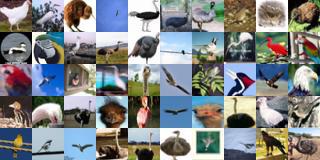}

    \end{minipage}
    \begin{minipage}[b]{0.5\textwidth}
     \centering
        \includegraphics[scale=0.6]{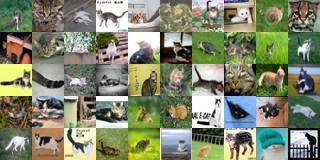}

    \end{minipage}
    
	\vspace{1.5ex}
    \begin{minipage}[b]{0.5\textwidth}
     \centering
        \includegraphics[scale=0.6]{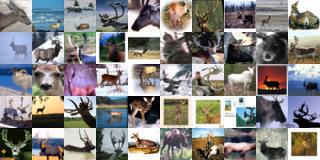}

    \end{minipage}
    \begin{minipage}[b]{0.5\textwidth}
     \centering
        \includegraphics[scale=0.6]{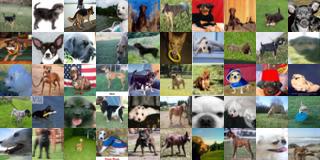}

    \end{minipage}
    
	\vspace{1.5ex}
    \begin{minipage}[b]{0.5\textwidth}
     \centering
        \includegraphics[scale=0.6]{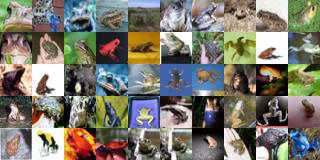}

    \end{minipage}
    \begin{minipage}[b]{0.5\textwidth}
     \centering
        \includegraphics[scale=0.6]{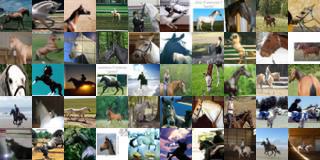}

    \end{minipage}
    
	\vspace{1.5ex}
    \begin{minipage}[b]{0.5\textwidth}
     \centering
        \includegraphics[scale=0.6]{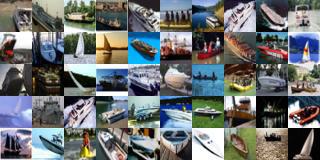}

    \end{minipage}
    \begin{minipage}[b]{0.5\textwidth}
     \centering
        \includegraphics[scale=0.6]{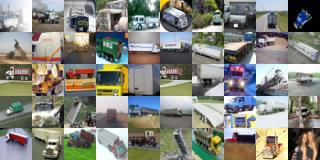}

    \end{minipage}
    \label{fig: imagenet-spurious2}
\end{figure}

\clearpage
\subsection{Visualization of the Searched ``Hard Examples (Loss)''}
\begin{figure}[htbp!]
    \caption{Visualization of the set searched by \methodls on CIFAR-10. We choose top 50 examples for each label.}
    \begin{minipage}[b]{0.5\textwidth}
     \centering
        \includegraphics[scale=0.6]{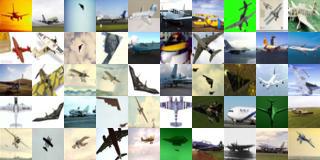}

    \end{minipage}
    \begin{minipage}[b]{0.5\textwidth}
     \centering
        \includegraphics[scale=0.6]{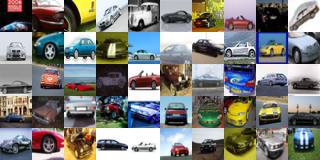}

    \end{minipage}

	\vspace{1.5ex}
    \begin{minipage}[b]{0.5\textwidth}
     \centering
        \includegraphics[scale=0.6]{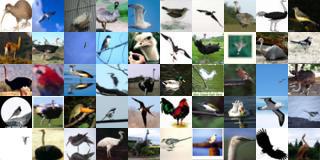}

    \end{minipage}
    \begin{minipage}[b]{0.5\textwidth}
     \centering
        \includegraphics[scale=0.6]{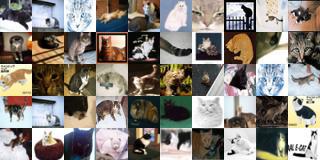}

    \end{minipage}
    
	\vspace{1.5ex}
    \begin{minipage}[b]{0.5\textwidth}
     \centering
        \includegraphics[scale=0.6]{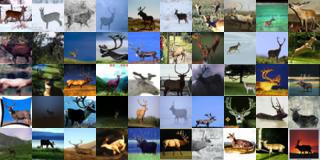}

    \end{minipage}
    \begin{minipage}[b]{0.5\textwidth}
     \centering
        \includegraphics[scale=0.6]{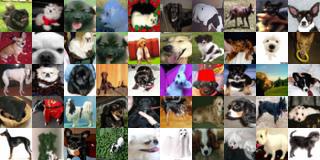}

    \end{minipage}
    
	\vspace{1.5ex}
    \begin{minipage}[b]{0.5\textwidth}
     \centering
        \includegraphics[scale=0.6]{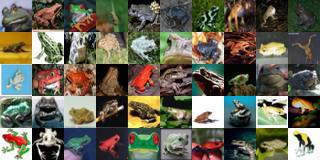}

    \end{minipage}
    \begin{minipage}[b]{0.5\textwidth}
     \centering
        \includegraphics[scale=0.6]{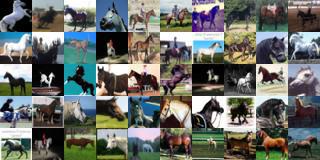}

    \end{minipage}
    
	\vspace{1.5ex}
    \begin{minipage}[b]{0.5\textwidth}
     \centering
        \includegraphics[scale=0.6]{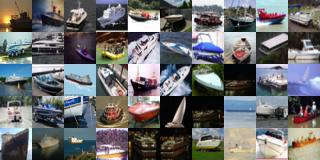}

    \end{minipage}
    \begin{minipage}[b]{0.5\textwidth}
     \centering
        \includegraphics[scale=0.6]{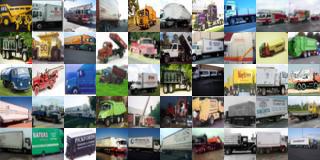}

    \end{minipage}
    \label{fig: imagenet-spurious2}
\end{figure}

\clearpage

\section{Sentences selected}

 \begin{table*}[htbp!]
	\centering
    \caption{Sentences of the set random selected on SST2. We sample randomly 10 examples for each label.}
    \footnotesize
 \resizebox{0.95\linewidth}{!}{%
		\begin{tabular}{|l|c|}
			\toprule
   \textbf{sentence} & \textbf{label}\\
   \midrule
inconsistent , meandering , and sometimes dry plot  & \multirow{10}{*}{negative}\\
made a great saturday night live sketch , but a great movie it is not  & \\
an mtv , sugar hysteria ,  & \\
was only  & \\
it 's been 13 months and 295 preview screenings since i last walked out on a movie , but resident evil really earned my indignant , preemptive departure  & \\
act weird  & \\
humbuggery ...  & \\
90 punitive minutes of eardrum-dicing gunplay , screeching-metal smashups , and flaccid odd-couple sniping .  & \\
of screenwriting cliches that sink it faster than a leaky freighter  & \\
sit still for two hours and change watching such a character , especially when rendered in as flat and impassive a manner as phoenix 's  & \\
   \midrule
a smart , solid , kinetically-charged spy flick worthy of a couple hours of summertime and a bucket of popcorn  & \multirow{10}{*}{positive}\\
great acting  & \\
have ever seen , constantly pulling the rug from underneath us , seeing things from new sides , plunging deeper , getting more intense  & \\
is a film in which the talent is undeniable  & \\
come away with a greater knowledge of the facts of cuban music  & \\
shows how deeply felt emotions can draw people together across the walls that might otherwise separate them .  & \\
the crazy things that keep people going in this crazy life  & \\
appeal to asian cult cinema fans and asiaphiles interested to see what all the fuss is about .  & \\
potentially interesting  & \\
thrusts the audience  & \\
\bottomrule
		\end{tabular}%
 }
	\label{tab:02_main_nlp}
\end{table*} 

 \begin{table*}[htbp!]
	\centering
    \caption{Sentences of the set searched by \methodgd for SST2. We choose top 10 examples for each label.}
    \footnotesize
		\begin{tabular}{|l|c|}
			\toprule
   \textbf{sentence} & \textbf{label}\\
   \midrule
is well below expectations .  & \multirow{10}{*}{negative}\\
make it sting  & \\
is well below expectations  & \\
huge sacrifice  & \\
best spent elsewhere  & \\
few ` cool ' actors  & \\
laughably  & \\
below is well below expectations .  & \\
spare dialogue  & \\
temperamental  & \\

\midrule

to winger fans who have missed her since 1995 's forget paris  & \multirow{10}{*}{positive}\\
rocky and  & \\
becomes compulsively watchable  & \\
particularly balk , who 's finally been given a part worthy of her considerable talents  & \\
balk , who 's finally been given a part worthy of her considerable talents  & \\
clearly a manipulative film  & \\
entertainingly nasty  & \\
busts out of its comfy little cell  & \\
fascinate me  & \\
rediscovers his passion in life  & \\
\bottomrule
		\end{tabular}%
	\label{tab:02_main_nlp}
\end{table*}

 \begin{table*}[htbp!]
	\centering
    \caption{Sentences of the set searched by \methodls for SST2. We choose top 10 examples for each label.}
    \footnotesize
		\begin{tabular}{|l|c|}
			\toprule
   \textbf{sentence} & \textbf{label}\\
   \midrule
a damn fine and a truly distinctive and a deeply pertinent film  & \multirow{10}{*}{negative}\\
provides an invaluable service  & \\
is an undeniably worthy and devastating experience  & \\
gain the unconditional love she seeks  & \\
unfolds as one of the most politically audacious films of recent decades from any country , but especially from france  & \\
self-deprecating , biting and witty feature  & \\
reasonably creative eighth-grader  & \\
chilling tale  & \\
noble end  & \\
from sharing the awe in which it holds itself  & \\
\midrule

fails to have a heart , mind or humor of its own  & \multirow{10}{*}{positive}\\
terminally bland ,  & \\
's not a brilliant piece of filmmaking  & \\
after next spreads them pretty thin  & \\
an admittedly middling film  & \\
the movie is silly beyond comprehension ,  & \\
just a bunch of good actors flailing around in a caper that 's neither original nor terribly funny  & \\
, incoherence and sub-sophomoric  & \\
he script is n't up to the level of the direction  & \\
an overcooked souffl  & \\
\bottomrule
		\end{tabular}%
	\label{tab:02_main_nlp}
\end{table*}




\end{document}